\documentclass[a4paper,fleqn]{cas-sc}
\usepackage[utf8]{inputenc}
\usepackage{algorithm}
\usepackage{pifont}
\usepackage{hyperref}
\usepackage{algorithmic}
\hypersetup{
    colorlinks=true,
    linkcolor=blue,
    filecolor=cyan,
    citecolor=blue,
    urlcolor=black
} 
\usepackage{flushend} % Balance references
\usepackage{amsmath}
\usepackage[authoryear,longnamesfirst]{natbib}

\def\ALC@uniqueAutorefName{\relax}

\def\tsc#1{\csdef{#1}{\textsc{\lowercase{#1}}\xspace}}
\tsc{WGM}
\tsc{QE}
\tsc{EP}
\tsc{PMS}
\tsc{BEC}
\tsc{DE}
\usepackage{setspace}
\begin{document}

%%%%%%%%%%%%%%%%%%%% TITLE PAGE %%%%%%%%%%%%%%%%%%%%%%%
% \begin{titlepage}
% \begin{center}
% \vspace*{10pt}
% \doublespacing
% \textbf{\Large GCM-Net: Graph-enhanced Cross-Modal Infusion with a Metaheuristic-Driven Network for Video Sentiment and Emotion Analysis }
% \vspace*{20pt}

% % Author names and affiliations
% Prasad Chaudhari$^{a}$ (ms2204101003@iiti.ac.in), Aman Kumar$^b$ (ee210002012@iiti.ac.in), \\ Chandravardhan Singh Raghaw$^a$ (phd2201101016@iiti.ac.in), Mohammad Zia
% Ur Rehman$^a$ (phd2101201005@iiti.ac.in), \\ Nagendra Kumar$^a$ (nagendra@iiti.ac.in) \\

% \hspace{1pt}

% \begin{flushleft}
% \small  
% $^a$ Department of Computer Science and Engineering, Indian Institute of Technology Indore, Khandwa Road, Simrol, Indore, 453552, Madhya
% Pradesh, India\\
% % $^c$ Full address of last author, including the country name  
% $^b$ Department of Electrical Engineering, Indian Institute of Technology Indore, Khandwa Road, Simrol, Indore, 453552, Madhya
% Pradesh, India\\

% \vspace{2cm}
% \normalsize
% \textbf{Corresponding Author:} \\
% Nagendra Kumar \\
% Department of Computer Science and Engineering, \\
% Indian Institute of Technology Indore, Indore 453552, India \\
% Tel: +91-7316603225 \\
% Email: nagendra@iiti.ac.in

% \end{flushleft}        
% \end{center}
% \end{titlepage}
%%%%%%%%%%%%%%%%%%%%%%%%%%%%%%%%%%%%%%%%%%%%%%%%%%%%%%%

% \shorttitle{GCM-Net: An Optimized Coupled Transformer-Convolution Network for Leukemia Detection}
\shortauthors{Chaudhari \textit{et~al.}}

\title{\textbf{\Large GCM-Net: Graph-enhanced Cross-Modal Infusion with a Metaheuristic-Driven Network for Video Sentiment and Emotion Analysis 
}}

%  \shorttitle{Graph-enhanced Cross Modal Infusion with Metaheuristic-Driven
% Network for Video Sentiment and Emotion Analysis}

% NTR 
\author[1]{Prasad Chaudhari}

\ead{ms2204101003@iiti.ac.in}
\affiliation[1]{organization={Department of Computer Science and Engineering, Indian Institute of Technology Indore},
    addressline={Khandwa Road, Simrol}, 
    city={Indore},
    postcode={453552}, 
    state={Madhya Pradesh},
    country={India}}

\affiliation[2]{organization={Department of Electrical Engineering, Indian Institute of Technology Indore},
    addressline={Khandwa Road, Simrol}, 
    city={Indore},
    postcode={453552}, 
    state={Madhya Pradesh},
    country={India}}

\author[2]{Aman Kumar}
\ead{ee210002012@iiti.ac.in}


\author[1]{Chandravardhan Singh Raghaw}
\ead{phd2201101016@iiti.ac.in}

\author[1]{Mohammad Zia Ur Rehman}
\ead{phd2101201005@iiti.ac.in}

\author[1]{Nagendra Kumar}%[
\cormark[1]
\ead{nagendra@iiti.ac.in}
\cortext[cor1]{Corresponding author: Nagendra Kumar}
% NTR

\begin{abstract}
Sentiment analysis and emotion recognition in videos are challenging tasks, given the diversity and complexity of the information conveyed in different modalities. Developing a highly competent framework that effectively addresses the distinct characteristics across various modalities is a primary concern in this domain. Previous studies on combined multimodal sentiment and emotion analysis often overlooked effective fusion for modality integration, intermodal-contextual congruity, optimizing concatenated feature spaces, leading to suboptimal architecture. This paper presents a novel framework that leverages the multi-modal contextual information from utterances and applies metaheuristic algorithms to learn the contributing features for utterance-level sentiment and emotion prediction. Our \textbf{G}raph-enhanced \textbf{C}ross-Modal Infusion with a \textbf{M}etaheuristic-Driven \textbf{Net}work (\textbf{GCM-Net}) integrates graph sampling and aggregation to recalibrate the modality features for video sentiment and emotion prediction. GCM-Net includes a cross-modal attention module determining intermodal interactions and utterance relevance. A harmonic optimization module employing a metaheuristic algorithm combines attended features, allowing for handling both single and multi-utterance inputs. To show the effectiveness of our approach, we have conducted extensive evaluations on three prominent multi-modal benchmark datasets, CMU MOSI, CMU MOSEI, and IEMOCAP. The experimental results demonstrate the efficacy of our proposed approach, showcasing accuracies of 91.56\% and 86.95\% for sentiment analysis on MOSI and MOSEI datasets. We have performed emotion analysis for the IEMOCAP dataset procuring an accuracy of 85.66\% which signifies substantial performance enhancements over existing methods. 
% These results highlight substantial performance enhancements over state-of-the-art methods on these datasets.
\end{abstract}
\maketitle
\thispagestyle{empty}
\section{\MakeUppercase{Introduction}}
\label{section:introduction}
The social media evolution driven by the widespread adoption of mobile and networking technology, has undergone a remarkable transformation marked by an unprecedented surge in multimodal content and a significant increase in the volume of expressions observed on these platforms. Users effortlessly utilize videos \citep{poria-etal-2017-context} to convey a diverse set of expressions, reflecting their sentiments and emotions through the integration of text, audio, and visual data. This shift towards multimodal content \citep{SHI2022655} underscores the need for refined methods and data analysis techniques to navigate the intricacies of human emotions embedded within multimedia content. Recognizing this evolving landscape, our research introduces a novel Video Sentiment Analysis and Emotion Recognition framework, as a solution to figure out the detailed reciprocation of human sentiments and emotions encapsulated in videos.

\subsection*{\textit{Advancing from Uni-Modality to Multi-Modality }}

Traditionally, sentiment analysis and emotion recognition tasks were predominantly centered around unimodal approaches \citep{10.1002/widm.1253}, with a primary focus on textual content where the inter-relationships among words and phrases were considered. As social media evolved to include more multimodal content, the shortcomings of unimodal approaches \citep{devlin2019bert} became more evident. Consequently, depending solely on textual content \citep{heirarchy-text} proves inadequate for extracting human sentiments, as the interpretation of speaker expressions often evolves dynamically, influenced by non-verbal behaviors~\citep{msa-fusion-survey}. This unimodal approach relying on text also posed challenges in understanding the intricate details conveyed through visual and audio data \citep{10.1145/3394171.3413690}, as well as need of efficient techniques for fusion of these modalities.

\begin{figure}
    \centering
    \includegraphics[width= 11 cm]{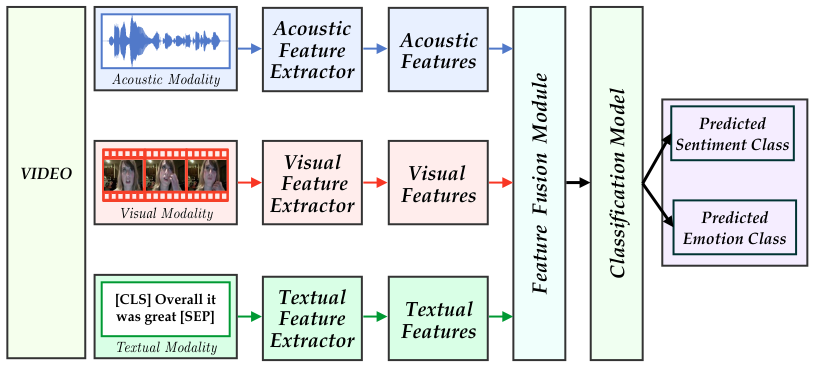}
    \caption{Generic Methodology Structure for Multimodal Sentiment Analysis and Emotion Prediction}
    \label{fig:GeneralTrend}
\end{figure}

The Video Sentiment Analysis and Emotion Recognition (VSAER) task aligns with the paradigm shift towards multimodality. It contrasts the traditional text-based approaches~\citep{tfn} by incorporating information from various modalities, such as visuals and acoustics. This procedure is illustrated in a generalized way in \autoref{fig:GeneralTrend}. However, this integration presents a significant challenge due to the inherent heterogeneity of sentiment and emotional information across these modalities. Unlike unimodal analysis, where each modality carries consistent semantic meaning, multimodal signals in VSAER are often disparate. Text, for example, is composed of discrete words with specific meanings, while visuals and audio consist of continuous digital signals. This disparity necessitates a robust fusion framework to effectively integrate these heterogeneous sentiment sources which turns out as a crucial aspect in VSAER research domain.

\subsection*{\textit{Research Trajectory in Video Sentiment Analysis and Emotion Recognition}}
% various techniques for video sentimental analysis

Prior research works have emphasized the development of comprehensive sentimental and emotional representations, capturing intricate intra-modal and inter-modal interactions through fusion techniques and yielded notable advancements in VSAER tasks. 
As shown in Figure~\ref{fig:GeneralTrend}, generic architectures utilize multi-input from a video and generate emotion or sentiment as output, where different fusion mechanisms are employed for cross-modal information extraction. 

Considering these approaches, more concentration was on diverse fusion architectures to facilitate interactions within and between modalities. These approaches can be broadly considered feature-fusion-based, graph-based, and deep learning-based models. We investigated feature-fusion-based models such as THMM (Tri-modal Hidden Markov Model) by \citep{10.1145/2070481.2070509} that extracts features from all modalities, concatenates them, and passes through a tri-modal HMM classifier. The TFN (Tensor Fusion Network) proposed by \citep{tfn} is based on the concept that learns both the intra-modality and inter-modality dynamics end-to-end using tensor fusion. The LMF (Low-rank Multimodal Fusion) \citep{lmf} model works by capturing the low-rank tensors for efficient multimodal representation. We then explored graph-based models \citep{heirarchy-text} such as Adversarial Representation Graph Fusion (ARGF) \citep{argf} uses adversarial learning to refine representations with graph fusion. Multi-channel Attentive Graph Convolutional Network (MAGCN) \citep{magcn} combines multi-channel attention with graph convolutions for joint learning. Deep Learning model, COSMIC \citep{cosmic} combines cross-modal similarities using a shared attention mechanism. The SIMR \citep{simr} learns shared multimodal representations while preserving modality-specific details. The AOBERT \citep{aobert} extends BERT for joint encoding of text and visual information.

% \textbf{Areas of Concern in Prior Research}
% Challenges or limitations of SOTA
% \subsection*{\textit{Unaddressed Aspects in Existing Research}}

\subsection*{\textit{Gaps in Existing Research}}

Multimodal sentiment analysis models both intra-modal dynamics and inter-modal dynamics. Intra-modality dynamics represent interactions within a specific modality, such as interactions between words in a sentence. Inter-modality dynamics denotes the interactions between different modalities. 
Current multimodal sentiment analysis methods struggle to concurrently quantify both intra-modal and inter-modal dynamics. To address this, we propose a novel technique that explicitly harnesses both types of dynamics. However, capturing these dynamics often presents a challenge with traditional fusion approaches. 

Further, previous approaches, while applying attention solely to the contextual utterance for classification did not fully account for the correlations among the modalities of the target utterance and the context utterances, which in turn hindered the accurate distinction of the most relevant modalities for sentiment and emotion \citep{LIU2023679} prediction of the target utterance. Consequently, this approach resulted in suboptimal multimodal feature representation when combining modalities from the context with those of the target utterance \citep{huang2024dynamic}. We therefore extend our research, by building upon the strengths and constructively focusing on the limitations identified in earlier studies. 

\subsection*{\textit{Technical Insights of Proposed Framework and Contributions}}
% https://www.sciencedirect.com/science/article/pii/S1361841523002475 (Paper outline)

This paper introduces \textbf{G}raph-enhanced \textbf{C}ross-Modal Infusion with a \textbf{M}etaheuristic-Driven \textbf{Net}work (\textbf{GCM-Net}) for VSAER tasks. 
GCM-Net employs four key modules for enhanced sentiment and emotion analysis: Graph-based Feature Recalibration and Enrichment (FRE), Intermodal Contextual Interaction Module (ICIM), Harmonic Optimization Algorithm (HOA), and a classifier module.
FRE analyzes each utterance by building a network. It leverages modality-specific graphs to capture both temporal context and feature relationships with nearby features within each modality. This graph-based feature enrichment surpasses individual features alone, enabling more accurate cross-modal representation. Furthermore, FRE reconstructs features by graphically sampling and combining feature \citep{ZHAO202273} relativity within the modality-specific graph network. By reconstructing these features, FRE encompasses a broader range of features surpassing the individual features that assist in improved inter-modality feature reconstruction.
Further, we incorporate ICIM, which creates an amplified and more informative cross-modal feature representation. It computes pairwise attention scores between modalities, learning each modality's contribution to the overall sentiment and emotion. This cross-modal attention mechanism allows ICIM to capture interactions that might be missed by analyzing modalities independently. Finally, HOA is a metaheuristic approach that actively explores the solution space and selects an optimal subset of features. This effectively addresses data redundancy, which is a known challenge in late fusion approaches. HOA reduces dimensionality by selectively discarding irrelevant features, resulting in a compact and informative feature set. These optimized features are subsequently fed into a classifier, leading to demonstrably improved classification performance in comparison to the existing approaches.

We evaluate our Model on two subtasks: Multimodal Sentiment Analysis (MSA) and Multimodal Emotion Recognition (MER). Three public datasets are used: CMU-MOSI \citep{zadeh2016mosi}, CMU-MOSEI \citep{bagher-zadeh-etal-2018-multimodal}, and IEMOCAP \citep{Busso2008IEMOCAPIE}. The experimental results demonstrate that our model outperforms the existing techniques. Furthermore, the ablation study and further analysis prove the effectiveness of different components of our proposed model.
% \newline
Our main contributions to this paper are summarized as follows:

\begin{itemize}
 \item  We propose a unified video sentiment and emotion analysis framework named Graph-enhanced Cross-Modal Infusion with a Metaheuristic Driven Network for Video Sentiment and Emotion Analysis (GCM-Net). 
 \item This novel framework integrates a graph-based modality-specific feature recalibration approach to capture intricate details often unexamined by approaches based on early fusion techniques.

 \item We incorporate a Intermodal Contextual Interaction Module to dynamically assign weights to each modality's representation based on its significance in the fusion process. This ensures each modality contributes most effectively in the model training phase.

 \item We employ a harmonic optimization algorithm, to efficiently identify the optimal feature subset using a population-based metaheuristic approach. This solves the data redundancy challenge in late fusion models.
 
 \item Extensive experimental analysis on three benchmark datasets demonstrates that our proposed mode, GCM-Net model effectively addresses limitations of previous work and exhibits improved efficiency and generalizability compared to existing approaches.
\end{itemize}

The organization of this paper is as follows: Section~\ref{section:related-works} provides an
overview of the related work, discussing the current state-of-the-art research on multimodal sentiment and emotion prediction. The problem formulation is explained in Section ~\ref{section:formulation}. Furthermore, in Section ~\ref{section:methodology} we illustrate our proposed approach and implementation details to cater the problem definition. The are provided in Section ~\ref{section:experimental-evaluation} presents experimental results conducted on public datasets  to show the performance and robustness of the proposed architecture. Finally, our proposed work is summarized with the conclusion in Section~\ref{section:conclusion}.

\section{\MakeUppercase{Related Works}}
\label{section:related-works}
This section provides a comprehensive review of existing research on VSAER. Recent studies have proposed various VSAER models, which can be broadly categorized into two main architectural approaches: Deep Learning-based Methods and Modality Fusion-based Methods with Graph-based models. We delve into each of these approaches in detail within the following subsections.

\paragraph{ \textbf{A) Deep Learning-based Methods}}
\subparagraph{}
This section explores deep learning approaches for sentiment or emotion analysis, particularly in the context of dialogue scenarios where sentiment interactions are often more complex. Capturing the subtle sentiment associations between participants in a conversation remains a significant challenge. Hazarika et al.~\citep{icon} proposed an interactive conversational memory network (ICMN) that leverages global memories to generate contextual summaries, facilitating multimodal sentiment detection. Zhang et al.~\citep{qlm} introduced a novel quantum-inspired interactive network (QLM) that combines elements of quantum theory with Long Short-Term Memory (LSTM) networks~\citep{rajagopalan2016extending} to capture both intra-utterance and inter-utterance interaction dynamics. Additionally, Ghosal et al.~\citep{cosmic} presented the COSMIC framework, which incorporates commonsense reasoning to learn the interrelationships between speakers in a conversation.

The emergence of deep learning technologies has significantly impacted various research fields due to their impressive performance. Among these, the Transformer model, renowned for its application in machine translation, has garnered considerable attention. This sequence-to-sequence architecture leverages solely attention mechanisms, eschewing recurrent and convolutional structures. Notably, the Transformer establishes associations between each element within a sequence during sequential data modeling and context mining, leading to superior accuracy, stability, and speed. Consequently, researchers are actively exploring its potential in diverse domains beyond machine translation.

The Transformer model has been applied to unimodal representation correlation in a multiple research works. To record the interactions between multimodal sequences at various time steps, Tsai et al.\citep{tsai-etal-2019-multimodal} used a Multimodal Transformer (MulT). The Transformer has additionally shown promise in merging unimodal features for emotion analysis. Rahman et al.~\citep{rahman2020integrating} used huge pre-trained Transformers for multimodal information integration, while Delbrouck et al.~\citep{delbrouck2020transformer} used a Transformer framework to combine several unimodal representations for multimodal emotion analysis(MER). 

Recent advancements include the work of \citep{simr} which suggested a Transformer-based multimodal encoding-decoding translation network that prioritizes textual data using a combined encoding-decoding strategy, is one example of recent advances. The Speaker-Independent Multimodal Representation (SIMR) framework was established by  \citep{simr} to minimize the impact of customized speech and visual elements. This approach employs a Cross-modal Transformer module to concurrently identify compatible and incompatible cross-modal interactions, splitting nonverbal inputs into style encoding and content representation. Additionally, All-modalities-in-one Bidirectional Encoder Representations from Transformers (AOBERT), a single-stream Transformer pre-trained on two tasks concurrently, was presented by \citep{aobert} to capture linkages and dependencies between modalities. When taken as a whole, this research demonstrates how promising the Transformer approach is as a basis for multimodal sentiment analysis.

Furthermore, to address challenges in multimodal emotion recognition (MER), Dai et al.~\citep{DBLP:journals/corr/abs-2009-09629} present the EmoEmbs model, which introduces a modality-transferable approach using emotion embeddings. This model learns mapping functions to translate pre-trained word embeddings into visual and auditory spaces, hence representing emotion categories for textual input. The model determines the representation distance for each modality between the goal emotions and the input sequence and then uses this distance to forecast outcomes. This approach's reliance on pre-trained word embeddings may result in suboptimal performance for non-textual modalities.

\paragraph{\textbf{B) Modality Fusion-based  Methods}}
\subparagraph{}

Modality fusion-based methods \citep{DAI2023164} for VSAER involves combining information from multiple modalities, such as audio, visual, and textual data, to achieve improved sentiment understanding. These methods can be broadly categorized into three main approaches: tensor-based fusion, translation-based fusion, and attention-based fusion.

Tensor-based fusion leverages tensors to represent and combine multimodal features. Representative models include the Tensor Fusion Network (TFN) \citep{tfn} and Low-rank Multimodal Fusion (LMF) \citep{lmf}. TFN employs a three-fold Cartesian product for multimodal representation fusion, while LMF focuses on optimizing fusion efficiency and temporal modeling. However, these methods may prioritize low-level features over contextual information, potentially hindering their performance in complex scenarios like dialogue analysis.

Translation-based fusion approaches multimodal sentiment analysis by translating representations from one modality to another. The Multimodal Cyclic Translation Network (MCTN) \citep{DBLP:journals/corr/abs-1812-07809} exemplifies this approach, where representations are iteratively translated between modalities to learn increasingly discriminative joint representations. While MCTN enhances robustness to missing or corrupted modalities, the cyclic translation process can be computationally expensive and time-consuming.

Attention-based fusion utilizes attention mechanisms to selectively focus on relevant information from each modality during the fusion process. Models like the Multi-attention Recurrent Network (MARN) \citep{marn}, Recurrent Attended Variation Embedding Network (RAVEN) \citeauthor{raven}, and modal-utterance-temporal attention (MUTA) \citep{TANG2023119125} fall under this category. Attention-based methods generally outperform other fusion approaches in sentiment analysis and emotion recognition tasks. However, their parallel structure might neglect the inherent coherence of human emotions, and simple concatenation techniques commonly employed may overlook modality-specific information.

Moreover, ~\citep{DBLP:journals/corr/abs-2103-09666} proposed a MESM that builds a fully end-to-end model that connects and jointly optimizes the two phases for the Multimodal emotion recognition (MER) task. Here they rearrange the current datasets to make end-to-end training easier. The feature extraction process made use of a sparse cross-modal attention mechanism in order to minimize the computing overhead caused by the end-to-end model. However, one shortcoming of this approach is that the rearrangement of datasets might limit its generalizability to other datasets or real-world scenarios.

\paragraph{ \textbf{C) Graph-based Methods}}
\subparagraph{}

In VSAER, graph neural networks (GNNs) have become an effective means for modeling intricate interactions between all three modalities. In contrast to conventional models, GNNs depict data as expressive graphs, in which nodes stand for distinct modalities and edges show how they interact. GNNs can discover hidden dependencies and acquire richer representations thanks to this graph-based method, which eventually improves sentiment prediction accuracy.

Several recent studies have exhibited the efficacy of GNNs by using a hierarchical graph neural network to build the encoded multimodal representation fusion, the Adversarial Representation Graph Fusion model (ARGF) \citep{argf} uses adversarial training to overcome the modality distribution gap. For learning in-depth intra and inter-modal temporal relationships, the multimodal graph network turns unaligned sequences into a graph and uses cutting-edge graph convolution and pooling methods. Moreover, the Multi-channel Attentive Graph Convolutional Network (MAGCN) \citep{magcn} integrates sentiment-related knowledge into inter-modality feature representations by utilizing multi-head self-attention and densely connected graph convolutional networks to capture inter-modality dynamics. Further, in addition to graph neural network and attention, our model effectively capitalizes the modality fusion, intermodal contextual congruity, and suboptimal feature space optimization that facilitates enhanced video understanding and classification.

% Furthermore, we will examine GCM-Net's architecture and its components that contribute to its superior performance. The study highlights the ability of the model to identify sentiment specifics across modalities.

\section{\MakeUppercase {Definition and Formulations}}
\label{section:formulation}
This study investigates the problem of automatically analyzing sentiment and emotion in video data. We consider a corpus of $M$ videos, denoted as $V = \{V_i\}_{i=1}^{M}
$. Each video is segmented into a sequence of $N$ utterances, represented as $U = \{U_i\}_{i=1}^{N}
$. For a given video $V_j \in V$, its corresponding utterance sequence is denoted as $U_j = \{U_{j,i}\}_{i=1}^{N}
$, where, $U_{j,i}$ refers to the $i^{th}$ utterance within that video.
Therefore, each utterance is represented as a multimodal sequence $U_{j,i}^{m}$, where, $U_{j,i}^{m} \in \{U_{j,i}^{t}, U_{j,i}^{a}, U_{j,i}^{v}\}$ corresponds to a raw unimodal sequence extracted from the $i^{th}$ utterance of the $j^{th}$ video.
\newline
Our research is focused towards achieving two primary objectives, each addressing key aspects of sentiment and emotion analysis within multimodal data. We aim to create a combined architecture that possesses capability to accurately predict both emotion and sentiment from the input data.
\subsubsection*{Problem 1: Sentiment Analysis}

The sentiment analysis task involves predicting the sentiment label $S_{i}^{s} \in \{0, 1\}$ for a given multimodal utterance $U_{j,i}^{m}$. Here, $1$ represents positive sentiment and $0$ represents negative sentiment. We aim to learn a function $f_S: U_{j,i}^{m} \mapsto S_{i}^{s}$ that effectively maps multimodal utterances to their corresponding sentiment categories.

\subsubsection*{Problem 2: Emotion Recognition}

We explore emotion prediction in addition to sentiment to extract more meaningful information from the data. Mathematically, this task seeks to predict the emotion category $S_{i}^{e} \in \{1, 2, \dots, C\}$ for a given multimodal utterance $U_{j,i}^{m}$, where, $C$ denotes the total number of emotion categories. We aim to learn a function $f_E: U_{j,i}^{m} \mapsto S_{i}^{e}$ that efficiently maps multimodal utterances to their corresponding emotion categories.

Our proposed model, denoted as \(\text{GCM-Net}\), intricately fuses information from text (\(U_{i}^{t}\)), acoustic (\(U_{i}^{a}\)), and visual (\(U_{i}^{v}\)) modalities to derive these predictions as illustrated in \autoref{eq:problemstatement}:

\begin{equation}
\label{eq:problemstatement}
S_{i}^{E}, S_{i}^{S} = \textbf{\text{GCM-Net}}(U_{i}^{t}, U_{i}^{a}, U_{i}^{v})
\end{equation}

This study presents an innovative and methodical approach that contributes significantly to the understanding of the intricate sentiment and emotion themes inherent in multimodal video data.

\begin{figure*}
    \centering
    \includegraphics[width=\textwidth]{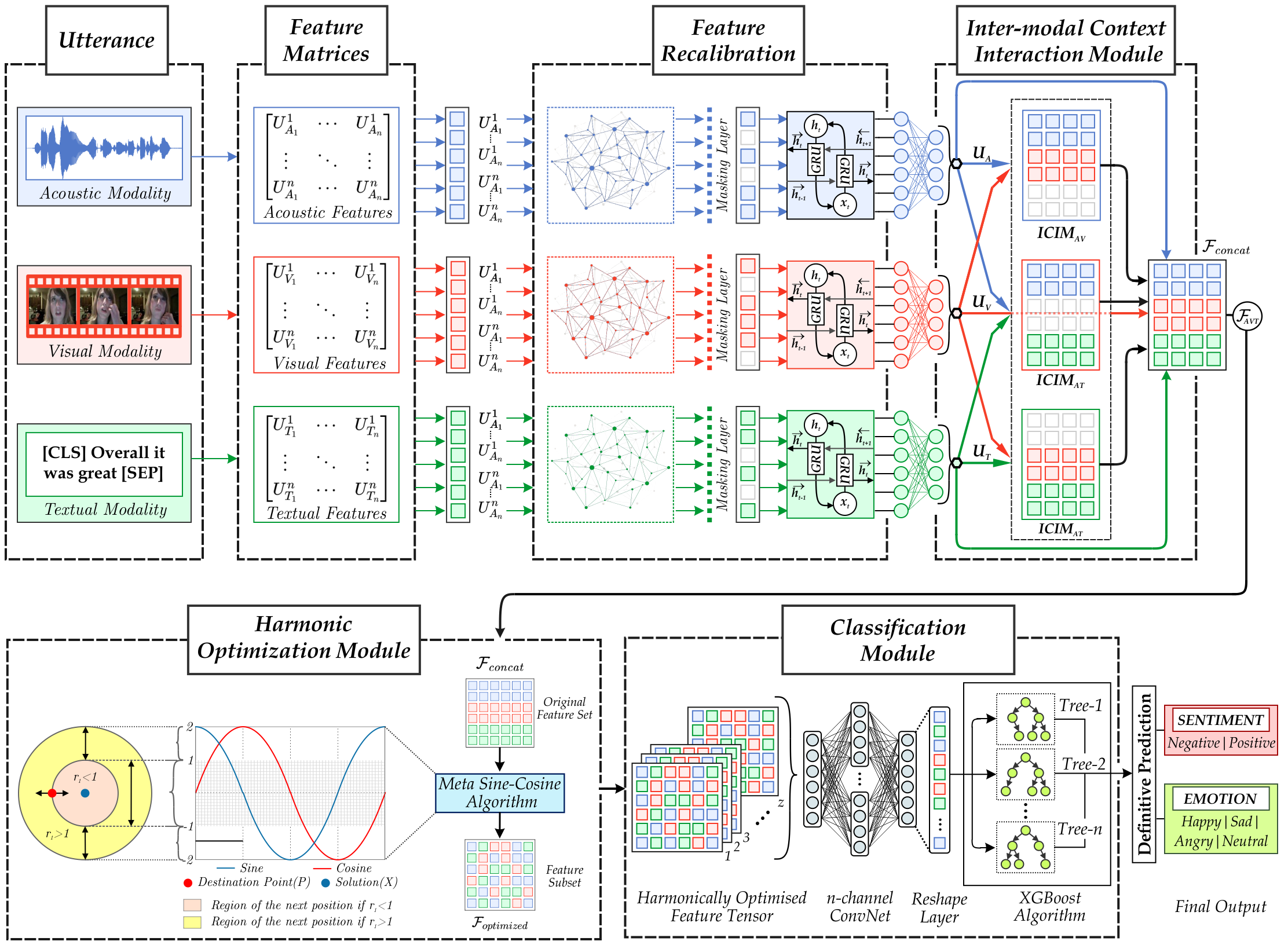}
    \caption{This figure illustrates our Graph-enhanced Cross-Modal Infusion (GCM-Net) architecture for video sentiment and emotion analysis. Modality-specific features are recalibrated via a graph-based approach, followed by dynamic weighting through ICIM. A harmonic optimization algorithm then selects optimal feature subsets for the ConvXGB classifier, achieving efficient and accurate prediction.}
    \label{fig : architecture}
\end{figure*}
\section{\MakeUppercase{Methodology}}
\label{section:methodology}
In this section, we elaborate on our architecture and the modality fusion approaches adopted for precise sentiment and emotion classification.
Our proposed approach, as illustrated in \autoref{fig : architecture}, employs distinct submodules to achieve multimodal sentiment analysis. First, we project the input multimodal video data into feature adjacency matrices. The Feature Recalibration and Enrichment (FRE) module then refines these matrices by leveraging graph sampling, aggregation, and Bi-GRU layers. Subsequently, the Intermodal Contextual Interaction Module (ICIM)  analyzes the interactions between all three modalities, fusing the enriched multisource representations. To address data redundancy, the Harmonic Optimization module selects an optimal feature subset based on calculated fitness. Finally, the classification module utilizes this refined feature subset to generate sentiment and emotion predictions.

\subsection{Feature Recalibration and Enrichment}
We take input videos from benchmark datasets, segment them into utterances, and obtain embeddings of textual, audio, and video data, respectively. Each video consists of a sequence of these elements, where, \(N\) is the number of segments in the utterance. As for every existing video, we have \(V = \{U_1, U_2, \ldots, U_N\}\), where each utterance is comprised of different elements as illustrated in~\autoref{eq:eq2}.
\begin{equation}
\label{eq:eq2}
 U_{i}^{m} = \{(U_{i}^{t}, U_{i}^{a}, U_{i}^{v})\}_{i=1}^N
\end{equation}
Here, $U_{i}^{t}, U_{i}^{a}, U_{i}^{v}$ denote the textual, acoustic and visual segments of utterance $U_{i}$.

\newpage
\subsubsection{Feature Recalibration using Graph Sampling}

\begin{figure*}
    \centering
    \includegraphics[width=0.9\linewidth]{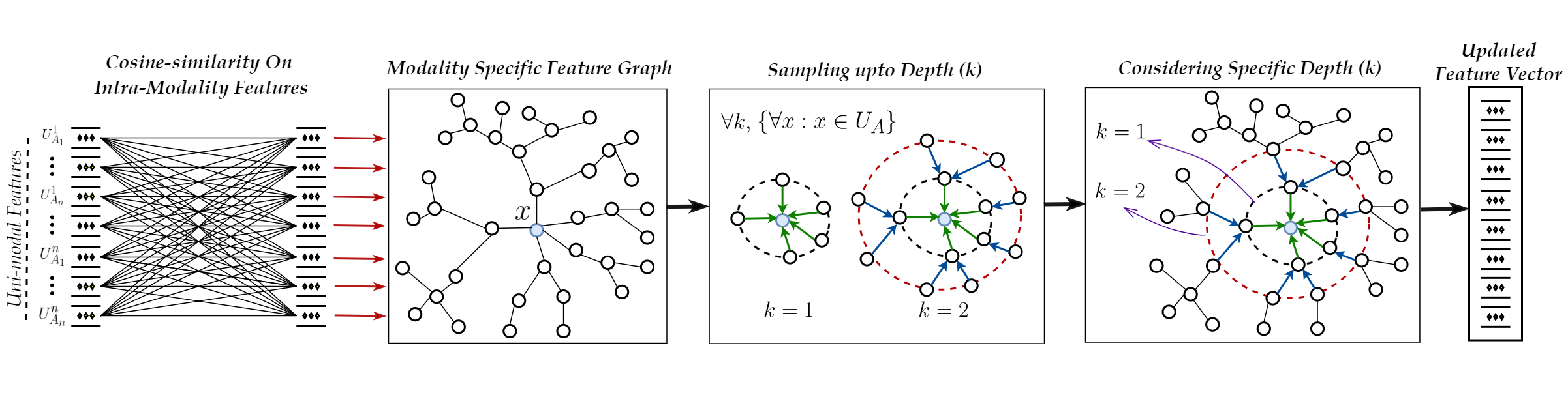}
    \caption{Illustration of Graph-based Feature Recalibration and Enrichment (FRE) which begins by constructing an adjacency matrix, linking utterances exceeding a similarity threshold. Neighboring features are then aggregated and sampled for node optimization}
    \label{fig:Graphsage}
\end{figure*}

For multimodal feature enrichment, the features extracted from each modality undergo a feature reconstruction process through a graph-based method. This process takes into account their temporal context and interrelation with neighboring features, measured by cosine similarity. This feature recalibration is performed independently for each modality as shown in \autoref{fig:Graphsage}.

\vspace{2mm}

\noindent\textbf{\textit{(a) Graph-based Sampling and Aggregation:}}
We use Graph Sampling and Aggregation (GraphSAGE), a variation of the Graph Convolutional Neural Network (GCN), for feature enrichment illustrated on lines 1-16 in Algoirthm~\ref{alg:algo1}. This method works by carefully selecting and compiling characteristics from a node's immediate vicinity inside the graph. GraphSAGE efficiently captures contextual information while enhancing computing efficiency by concentrating on the near neighborhood of each node, which is especially useful for large-scale graphs.

We initiate the process by calculating the cosine similarity-based adjacency matrix  \(A\) to capture relationships between feature embeddings and create an updated matrix as illustrated in \autoref{eq:adjacency_matrix}:

\begin{equation}
\label{eq:adjacency_matrix}
A_{ij} = 
\begin{cases}
1 & \text{if } similarity(U_i, U_j) \geq K_{threshold} \\
0 & \text{otherwise}
\end{cases}
\end{equation}

where, \(similarity(U_i, U_j) = cos(U_i, U_j)\) is the cosine similarity between embeddings \(U_i\) and \(U_j\).
This step selectively connects nodes exceeding a predefined similarity threshold, denoted as ($K_{threshold}$). This ensures that only highly similar nodes are linked with each other. The optimal value of $K_{threshold}$ is empirically chosen by performing multiple experiments.

\noindent\textbf{\textit{(b) Graph-based Loss Function:}}
The graph-based loss function illustrated in \autoref{eq:graphloss} promotes similar representations for nearby nodes in the graph represented by adjacency matrix \(A\) while enforcing distinct representations for dissimilar nodes:

\begin{equation}
\label{eq:graphloss}
J_G(z_u) = -\log(\sigma(z_u^T z_v)) - Q \cdot {E}_{n \sim P_n(v)} \log(\sigma(-z_u^T z_{vn}))
\end{equation}

In the above loss function, \(v\) is a node co-occurring near \(u\) in a fixed-length random walk. The element \(E_{n \sim P_n(v)}\) represents expectation over negative samples, where, \(n\) is sampled from the negative sampling distribution \(P_n(v)\). Furthermore, \(Q\) defines the number of negative samples. Finally, \(\log(\sigma(-z_u^T z_{vn}))\), measures the dissimilarity between the representation of node \(u\) and the representations of the negatively sampled nodes \(vn\).
\vspace{0.5\baselineskip}

\noindent\textbf{\textit{(c) Aggregator Function:}}
 The model incorporates an aggregator function to combine the contextual information extracted from the Bi-GRU layer for each segment or utterance. While various options exist, such as mean, LSTM, and pooling aggregators, extensive evaluations revealed that the LSTM aggregator consistently outperformed the others. Consequently, the LSTM aggregator is employed within the model.
% \textbf{LSTM Aggregator (\(\mathfrak{L}\)):}

\textbf{\textit{LSTM Aggregator:}}
The LSTM aggregator is based on an LSTM architecture, capturing long-term dependencies and intricate temporal patterns. It is defined in line 4 in Algorithm~\ref{alg:algo1}:
\begin{align}
\label{eq:lstmaggregator}
    h_k^v &= \sigma(\mathbf{W}_k \cdot \text{LSTM}(\mathbf{h}_{k-1}^N) + b_k)
\end{align}

As shown in~\autoref{eq:lstmaggregator}, 
the element $\mathbf{h}_{k-1}^N$, can be explained as, $\mathbf{h}_{k-1}^N = \text{CONCAT}(h_{k-1}^{u_1}, h_{k-1}^{u_2}, \ldots, h_{k-1}^{u_N})$ and $b_k$ is a bias term.

\vspace{2mm}

\noindent\textbf{\textit{(d) Graph-based Recalibrated Feature:}}
The final enriched features for each segment $v$ are obtained using a transformation function as illustrated in \autoref{eq:graphsage-feature}:
\begin{align}
\label{eq:graphsage-feature}
 \boldsymbol{\mathcal{G}}_{\text{v}} = \sigma(\mathbf{W}_T h_K^v + b_T)
\end{align}

where, the variable $\mathbf{W}_T$ donotes the learnable weight matrix and  $b_T$ signifies the corresponding bias term.

\subsubsection{Bi-directional and Contextual Feature Enrichment}
\label{subsec:bi-directional-enrichment}

To enhance the procured multimodal features, we integrate a Bidirectional Gated Recurrent Unit  (Bi-GRU) \citep{bigru} layer into our model architecture. This critical addition of Bi-GRU layer empowers our model to capture temporal dependencies inherent in the data, enabling a profound understanding of sequential patterns within multimodal features. For each segment or utterance \(v\), this integrated Bi-GRU layer conscientiously computes both forward (\(\overrightarrow{\mathbf{h}}_v\)) and backward (\(\overleftarrow{\mathbf{h}}_v\)) hidden states. These hidden states further assist in the forward pass propagation.
\vspace{2mm}

\noindent\textbf{\textit{(a) Forward Pass (Forward Hidden States):}}
In the Bi-GRU layer, the forward hidden states \(\overrightarrow{a}_t\) are iteratively computed for each time step (t). This calculation involves applying an activation function \(g_1\) to a weighted combination of the previous forward hidden state \(\overrightarrow{a}_{t+1}\), the current input element $x_t$, and learnable weight matrices (\(W_{aa}\) and \(W_{ax}\)).

\vspace{2mm}

\noindent\textbf{\textit{(b) Backward Pass (Backward Hidden States):}}
Similarly, the backward hidden states (\(\overleftarrow{a}_t\)) are calculated for each time step in the backward direction. The process mirrors the forward pass, using the same activation function \(g_1\) but considering the previous backward hidden state (\(\overleftarrow{a}_{t-1}\)) instead of the forward hidden state.
\vspace{2mm}

\noindent\textbf{\textit{(c) Final Hidden States (Combining Forward and Backward States):}}
Once both forward and backward hidden states are obtained, they are concatenated to form the final hidden states (\(a_t\)) for each segment \(v\). This final representation incorporates contextual information from both directions, enabling the Bi-GRU to capture a more comprehensive context for downstream tasks.

\subsubsection{Dimension Consistency via Multimodal Feature Projection to Dense Layers}

Three distinct Bidirectional Gated Recurrent Unit (Bi-GRU) layers are successively processed through the output from the graph recalibration process ({$\boldsymbol{\mathcal{G}}_v$}), using forward and backward state concatenation. As a result, fully connected dense layers get the outputs from the Bi-GRU layers, reducing the dimensionality of the features to a common size. Three matrices are produced as a result: 
\(T_{bigru} \in \mathbb{R}^{u \times d}\) (text), \(V_{bigru} \in \mathbb{R}^{u \times d}\) (visual), and \(A_{bigru} \in \mathbb{R}^{u \times d}\) (acoustic), where, \(u\) represents the number of utterances and \(d\) is the number of neurons in the dense layer.

\subsection{Intermodal Contextual Interaction Module (ICIM)}
\label{subsec:cross-modal-attention}
\begin{figure}
    \centering
    \includegraphics[width= 9cm]{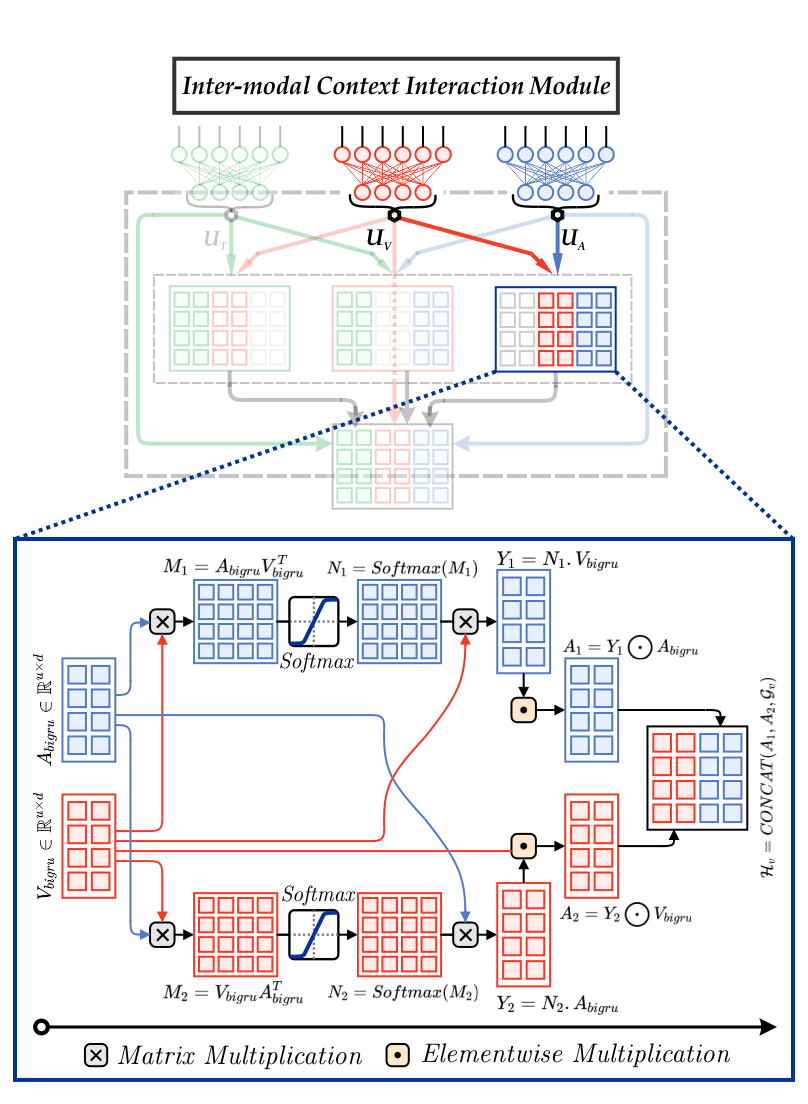}
    \caption{Intermodal Contextual Interaction Module (ICIM). This module facilitates cross-modal interaction by computing pairwise attentions between different modalities}
    \label{fig:ICIM}
\end{figure}

The modality-specific features are procured from the dense layer channels for each audio, video, and textual data with uniform dimensionality. 
They are fed forward to the ICIM which is based on a cross-modal attention mechanism aiming to capture the interactions between different modalities illustrated in \autoref{fig:ICIM}. We compute pairwise attentions for each pair of modalities such as ($V_{bigru}$ \& $T_{bigru}$), $T_{bigru}$ \& $A_{bigru}$), and ($A_{bigru}$ \& $V_{bigru}$).
As shown in~\autoref{eq:attnscores}, we can calculate the attention scores between queries and keys :
\begin{equation}
\label{eq:attnscores}
a_{ij} = \text{softmax} \left( \frac{R_i \cdot K_j^T}{\sqrt{d_k}} \right)
\end{equation}
where, $a_{ij}$ is the attention score between the $i$-th token in the query and the $j$-th token in the key, $R$ is the query matrix, $K$ is the key matrix and $d_k$ is the dimensionality of the keys.

In particular, for ($A_{bigru}$ \& $V_{bigru}$), we obtain the modality representations of $A_{bigru}$ and $V_{bigru}$ from the Bi-GRU network, which encodes the contextual information of the utterances for each modality coming from the graph enrichment process. We then compute a pair of matching matrices \(M_1, M_2 \in \mathbb{R}^{u \times u}\) over the two representations, which measure the cross-modal similarity between the utterances as illustrated in~\autoref{eq:bi-modal-attention}.
\begin{equation}
\label{eq:bi-modal-attention}
M_1 = A_{bigru}V_{bigru}^T \quad \text{and} \quad M_2 = V_{bigru}A_{bigru}^T
\end{equation}
The probability distribution scores $N_1$ and $N_2$ are computed over $M_1$ and $M_2$ using the softmax function 
 as shown in~\autoref{eq:y1y2} to compute modality-wise attentive representations ($Y_1$ and $Y_2$):
\begin{align}
\label{eq:y1y2}
    Y_1 &= N_1.V_{bigru} \quad \text{and} \quad  Y_2 = N_2.A_{bigru}
\end{align}
Finally, the attention matrices $A_1$ and $A_2$ is obtained by taking dot product again with V and A respectively.
\begin{align}
    \label{eq:attention_matrix}
    A_1 &= Y_1 \odot A_{bigru} \quad \text{and} \quad A_2 = Y_2 \odot V_{bigru}
\end{align}
These attention matrices $A_1$ and $A_2$ computed in \autoref{eq:attention_matrix} are then concatenated to original refined values to increase feature size and the final refined embeddings are mentioned to be \textbf{$H_v$}.

\subsection{Feature Optimization Techniques}

This section explains the functioning of the population-based metaheuristic algorithm for feature selection and optimization in the proposed model. Population-based optimization techniques, known for their randomized search for optimization problems, lack certainty in discovering a solution in a single execution. However, with increased optimization iterations and the number of random solutions, the likelihood of identifying the global optimum improves. Additionally, a local search technique called Adaptive $\beta$-Hill Climbing (A$\beta$HC) \citep{adaptive-beta} is employed to refine the acquired feature subset. The optimized feature set establishes a mapping between the feature set and output classes, facilitating the Conv-XGB Deep Learning Classifier in making the final classification leveraging the optimized feature set as input.

Building upon these embeddings, advanced optimization techniques were employed to enhance feature subsets. The Harmonic Optimization Algorithm (HOA) systematically explored the solution space, ensuring the selection of informative and contextually relevant features. This enriched feature set undergoes further refinement through the (A$\beta$HC) local search strategy, which assists in fine-tuning the selected features for optimal precision. The classification phase involves the utilization of the K-Nearest Neighbors (KNN) classifier \citep{knn-classifier}, providing distinctive insights into the complexities of the incorporated multimodal data.

\begin{figure*} % Full-width Algorithm
\centering{
\begin{minipage}{0.9\linewidth}
\begin{algorithm}[H]
    \caption{\textsc{Feature Recalibration and Optimization Algorithm}}
    % \label{alg:gcmnet}
    \label{alg:algo1}
    \begin{algorithmic}
        \vspace{1mm}
        \STATE
        \textbf{Input:} Videos divided into utterances for three modalities text, audio, and images $\textbf{U}_v^t$, $\textbf{U}_v^a$, $\textbf{U}_v^i 
    \forall v \in V$ \\
        \textbf{Output:} Optimized Features $\boldsymbol{\mathcal{F}}_{\text{v}}$ for each utterance \\
    \end{algorithmic}
    {\vspace{1mm}}
        {\textbf{Function:} Graph-Reconstruct(\textit{Graph} $\mathcal{G}$, \textit{Features} $\{\textbf{U}_v, \forall v \in V\}$, \textit{Depth} $K$, \textit{Weights} $\boldsymbol{\langle \mathbf{W}_k \rangle}$, \textit{Non-linearity} $\sigma$, \texttt{AGGREGATE}\(_k, \forall k \in \{1, ..., K\}\), $N : v \rightarrow 2^V$)
}
    \begin{algorithmic}[1]
        \vspace{1mm}
        \STATE Initialize node embeddings. $\mathbf{h}_0^v \leftarrow \textbf{U}_v, \forall v \in V$ 
        \FOR{$v \in V$}
            \FOR{$k = 1$ \textbf{to} $K$}
                \STATE Aggregate neighbors' information using $\texttt{AGGREGATE}_k$  \\ $\mathbf{h}_{k, N(v)} \leftarrow \texttt{AGGREGATE}_k(\{\mathbf{h}_{k-1}^u, \forall u \in N(v)\})$ 
                \STATE Update node embeddings with non-linearity \\  $\mathbf{h}_{k}^v \leftarrow \sigma(\boldsymbol{\mathbf{W}_k} \cdot \texttt{CONCAT}(\mathbf{h}_{k-1}^v, \mathbf{h}_{k, N(v)}))$
                \STATE L2 normalize node embeddings across all nodes \\ $\mathbf{h}_k^v \leftarrow \frac{\mathbf{h}_k^v}{\|\mathbf{h}_k^v\|_2}, \forall v \in V$
            \ENDFOR
            \STATE Return final vector representations for all nodes. $\boldsymbol{\mathcal{G}_v} \leftarrow \mathbf{h}_K^v$ 
            \STATE Apply BiGRU to capture temporal dependencies
            \FOR{each pair of modalities \(X\) and \(Y\)}
                \STATE Calculate attention scores \(A_{\text{XY}}\) and update modality \(X\) using \(Y\) and vice versa using it
                \STATE Compute matching matrices \(M_1\) and \(M_2\) using Eq. \eqref{eq:bi-modal-attention}
            \ENDFOR
            \STATE CONCAT these scores to $\boldsymbol{\mathcal{G}_v}$ to get $\boldsymbol{\mathcal{H}_v}$
        \ENDFOR
        \STATE \textbf{return} Enhanced Features $\boldsymbol{\mathcal{H}_v}$ \\
        {{\vspace{1mm}}
         \textbf{Function:} HOA $\mathcal{A}\beta\mathcal{HC}$(\textit{Enhanced Features} $\{\boldsymbol{\mathcal{H}_v}, \forall v \in V\}$, $\boldsymbol{\mathbf{W}_k}$, $\beta_{\text{hc}}^{\text{min}}$, $\beta_{\text{hc}}^{\text{max}}$, $T_{\text{max}}$, $P$)}
    % \end{algorithmic}
    \vspace{1mm}
    % \begin{algorithmic}[1]
    \vspace{1mm}
    \STATE Initialize solutions with random feature subsets
    \FOR{$t = 1$ \textbf{to} $T_{\text{max}}$}
        \STATE Generate uniformly distributed random numbers $r_1$, $r_2$, $r_3$, $r_4$ in the range $[0,1]$
        \FOR{each solution $i$}
            \STATE Calculate the exploration-exploitation factor: \\
            $\text{exploration factor} = r_{1,j}^{t} \times \sin{(r_{2,j}^{t})} \quad \text{if } r_{4,j}^{t} < 0.5 $ \\            $ \text{exploitation factor} = r_{1,j}^{t} \times \cos{(r_{2,j}^{t})} \quad \text{otherwise} $
            
            \STATE Update solution position using SCA equation:
            \[ \mathcal{P}_{i, j}^{t+1} = \mathcal{P}_{i, j}^{t} + \text{factor} \times |r_{3,j}^{t}D_j^t -\mathcal{P}_{i, j}^{t}| \]

            \STATE Randomly select a feature subset using $x_{\text{rand}}$
            \STATE Apply $\mathcal{A}\beta\mathcal{HC}$ local search for fine-tuning as shown in Eq. \eqref{eq:abhcn}, Eq. \eqref{eq:abhcb}
        \ENDFOR
    \ENDFOR
    \STATE\textbf{return} Optimized features $\boldsymbol{\mathbf{F}_v}$
\end{algorithmic}
\end{algorithm}
\end{minipage}}
\end{figure*}

\subsubsection{Harmonic Optimization Algorithm (HOA) for Feature Selection}

The Harmonic Optimization Algorithm (HOA) is a population-based metaheuristic algorithm inspired by sine and cosine trigonometric functions as shown in \autoref{fig : architecture}. In the context of multimodal sentiment analysis, HOA is applied for feature selection, efficiently exploring the solution space to choose an optimal subset of features for enhanced classification performance.

For feature selection and optimization, we use the population-based metaheuristic Harmonic Optimization Algorithm (HOA). HOA explores the solution space and identifies the ideal destination space by using population-based, iterative stochastic techniques that resemble the harmonic behavior of sine and cosine functions. The two main phases of HOA are usually exploration and exploitation illustrated in Algorithm~\autoref{alg:algo1}. In the exploration phase, regions of the search space that show promise are found by combining random solutions with a higher degree of randomization. To lessen the randomness in the variations, on the other hand, the exploitation phase gradually modifies the answers.

To initiate the optimization procedure using HOA, the search element adjusts its self-position based on the sine and cosine functions, as given in~\autoref{eq:hoaposition}.

 \begin{equation}
\label{eq:hoaposition}
   \mathcal{P}_{i, j}^{t+1} = \!
   \begin{cases}
    \!\mathcal{P}_{i, j}^{t} + r_{1,j}^{t} \times \sin{(r_{2,j}^{t})} \times |r_{3,j}^{t}D_j^t -\mathcal{P}_{i, j}^{t}|, r_{4,j}^{t} < 0.5 \\
    
    \!\mathcal{P}_{i, j}^{t} + r_{1,j}^{t} \times \cos{(r_{2,j}^{t})} \times |r_{3,j}^{t}D_j^t -\mathcal{P}_{i, j}^{t}|, r_{4,j}^{t} \geq 0.5 \end{cases}
\end{equation} 

Here, $\mathcal{P}_{i, j}^{t}$ denotes the position of the current solution in the $j^{th}$ dimension of the $i^{th}$ search element at the $t^{th}$ iteration. $r_{2,j}^{t}$, $r_{3,j}^{t}$, and $r_{4,j}^{t}$ are uniformly distributed random numbers and $D_j^t$ represents the position of the $j^{th}$ dimension of the destination point (best solution) at the $t^{th}$ iteration. A random number $r_{1,j}^{t}$ facilitates the transition from exploration to exploitation of the search space, determined by \autoref{eq:hoarandom}.

 \begin{equation}
\label{eq:hoarandom} 
    r_{1,j}^{t} = \alpha - t\frac{\alpha}{T}
\end{equation} 

Here, $\alpha$, $t$, and $T$ represent the constant number, the $t^{th}$ iteration, and the total number of iterations, respectively.

The value of $r_{1,j}^{t}$ decides whether the search area is for exploitation (destination solution region) $(r_{1,j}^{t} \in [-1,1])$ or exploration (feasible solution region) $(r_{1,j}^{t} \in [-1,-2]$ or $r_{1,j}^{t} \in [1,2])$ is mentioned on line 21 in Algorithm~\ref{alg:algo1}. The stochastic variable $(r_{2,j}^{t}$ defines the search agent's movement toward or away from the destination point, bounded within $[0,2\pi]$, in sync with a complete cycle of sine and cosine functions. $(r_{3,j}^{t}$ balances the exploration and exploitation rates by introducing a random weight between $(0,2)$. Furthermore, $r_{3,j}^{t}$ introduces a stochastic step size for the destination point, emphasizing $(r_{3,j}^{t}$>$1)$ or not emphasizing $(r_{3,j}^{t} $<$ 1)$ its impact. Finally, the parameter $r_{4,j}^{t}$ evenly transitions between the sine and cosine components, as given in \autoref{eq:hoaposition}.
% \paragraph{\textbf{Algorithm Overview: }}

As shown in lines 17-26 of Algorithm~\ref{alg:algo1}, HOA initializes a set of random solutions representing feature subsets. Through iterative evaluations using the objective function, HOA refines these solutions by smoothly transitioning between the exploration and exploitation phases. The algorithm employs sine and cosine functions to update solution positions and efficiently explore the search space.

\paragraph{\textbf{Solution Update Procedure: }}

The positions of destination points \(x_i^{t+1}\) are updated using the following equations, where, \(r_1\), \(r_2\), \(r_3\), and \(r_4\) are random numbers in the range \([0, 1]\) as illsutrated in \autoref{eq:soln-update-equation}:

\begin{equation}
\label{eq:soln-update-equation}
    x_i^{t+1} = x_i^t + r_1 \times \sin(r_2 \times \arcsin(r_3)) \times x_{\text{rand}}
\end{equation}

Here, \(x_{\text{rand}}\) is a random binary vector, indicating whether a feature is selected (\(1\)) or not (\(0\)). 

\subsubsection{Local Search: Adaptive Beta Hill Climbing}
After identifying the most efficient features through the metaheuristic algorithm Harmonic Optimization Algorithm (HOA), further enhancement of exploitation ability can be achieved by integrating the local search technique named Adaptive $\beta$-Hill Climbing (A$\beta$HC). A$\beta$HC is a feature optimization algorithm utilizing local search-based techniques. These search techniques are guided by a pair of control parameters $\mathcal{N}_{\textsc{hc}}$ and $\beta_{\textsc{hc}}$, respectively. By adjusting these parameters, the search technique finds the optimal trade-off between exploitation and exploration. Fine-tuning these parameters plays a significant role in optimization because it helps enhance the convergence rate. The parameter $\mathcal{N}_{\textsc{hc}}$ is initially set to a value close to $1$, but it gradually decreases as the search process iterates. This allows the algorithm to dynamically adjust $\mathcal{N}_{\textsc{hc}}$ to improve search performance, as given in~\autoref{eq:abhcn}.

 \begin{equation}
\label{eq:abhcn} 
    \mathcal{N}_{\textsc{hc}}^t = 1 - \frac{t^{\frac{1}{P}}}{T_{max}^{\frac{1}{P}}}
\end{equation} 

Here, $\mathcal{N}_{\textsc{hc}}^t$ represents the value of $\mathcal{N}_{\textsc{hc}}$ at time $t$, $P$ is a constant used to linearly decrease the value of $\mathcal{N}_{\textsc{hc}}$ to a value close to $0$, and $T_{max}$ represents the upper limit of iterations for A$\beta$HC algorithm.

Moreover, the $\beta$ parameter undergoes deterministic adaptation within a defined range $\in [\beta_{\textsc{hc}}^{min},\,\beta_{\textsc{hc}}^{max}]$, mathematically expressed in~\autoref{eq:abhcb}.

\begin{equation}
\label{eq:abhcb} 
    \beta_{\textsc{hc}}^t = \beta_{\textsc{hc}}^{min} + t \times \frac{\beta_{\textsc{hc}}^{max} - \beta_{\textsc{hc}}^{min}}{T_{max}}
\end{equation} 

Here, $\beta_{\textsc{hc}}^t$ denotes the rate of $\beta_{\textsc{hc}}$ at time $t$, $\beta_{\textsc{hc}}^{min}$ and $\beta_{\textsc{hc}}^{max}$ represent the minimum and maximum values of $\beta_{\textsc{hc}}$ respectively, $T_{max}$ is the total number of iterations, and $t$ signifies the current time. 

After applying A$\beta$HC based HOA to the graph enriched feature vector $\boldsymbol{\mathcal{H}}_v$ this function returns a masking array stating the selected features for final feature leaning and sentiment prediction given by $\boldsymbol{F}_v$.

\subsection{Feature Learning with Bidirectional Convolutional Processing}

The final feature learning is a bidirectional approach for sentiment prediction as mentioned in Algorithm~\ref{alg:algo2} having multiple stages to it. The first stage is the input layer. This is followed by convolutional layers, responsible for feature learning by applying convolution and bias to input features. Next, the reshape layer is integrated, and finally, the class prediction layer. One portion of each layer can be used for feature learning, and the other part can be used for class prediction shown in \autoref{fig : architecture}. Further, we explore the convolutional layers responsible for feature extraction in the following sections.

\textbf{\textit{a) Input Representation and Initial Processing:}}
The optimized embeddings, $F_v$ obtained after applying A$\beta$HC based HOA, serve as input to the model, as shown in line 1 of  Algorithm~\ref{alg:algo1}. This layer will directly pass onto the next convolutional layers for further feature extraction.

\textbf{\textit{b) Convolutional Feature Extraction:}}
This layer employs convolution and pooling operations to extract hierarchical features, capturing both simple textures and complex structures crucial for understanding multimodal interactions. We denote these convolutional network parameters as $\Theta_{\text{conv}}$, and perform the following operations as illustrated below.

{\textit{ Convolutional Operations:}} This operation applies filters of size \(K\) to the input sequences given on line 8 in Algorithm~\ref{alg:algo1}, sliding them across the sequence and capturing local patterns. Mathematically, this can be expressed in \autoref{eq:convolution}:
\begin{equation}
\label{eq:convolution}
    C_{v1, i} = \text{ReLU}\left(\sum_{k=1}^{K} F_{v, i+k-1} \cdot \Theta_{\text{conv}, k}\right)
\end{equation}
        
where, \(C_{v1, i}\) denotes the \(i\)-th feature map at the first convolutional layer for utterance \(v\), \(F_{v}\) represents the optimized embedding for utterance \(v\), \(\Theta_{\text{conv}, k}\) is the \(k\)-th convolutional filter and \(K\) is the filter size.\\
{\textit{Pooling Operations: }} This layer performs downsampling on the feature maps by selecting the maximum value within a predefined window size. This operation emphasizes the most prominent features within the local receptive field, reducing the spatial dimensionality of the data while potentially preserving essential information as illustrated in line 7 of the Algorithm~\ref{alg:algo2}. The specific implementation involves selecting the maximum value from each element within the window, resulting in a compressed representation of the input.

The last layer in the CNN stack is the reshape layer. Its input is the output of the pooling layer that came before it, usually flattened into a single-column vector illustrated in line 8 in Algorithm~\ref{alg:algo2}. The retrieved features are compressed and forwarded to the reshape layer when using this vector for classification or regression tasks. The reshape layer creates the network's prediction by learning intricate correlations between the characteristics and the intended output using a set of weights and biases.
\subsection{Classification}
 The features extracted from the prior CNN layer are first forwarded to a reshape layer and are further passed to the classification stage where we integrate the ConvXGB \citep{THONGSUWAN2021522} classifier. It leverages the XGBoost algorithm for sentiment and emotion classification as illustrated in Algorithm~\ref{alg:algo2}. This procedure is sequentially elaborated as follows.
\begin{algorithm}
    \caption{\textsc{Enhanced ConvXGB for Sentiment and Emotion Classification}}
    \label{alg:algo2} 
    
        \textbf{Input:} Training dataset $D_{\text{Tr}} = \{(\mathbf{F}_v, \mathbf{S}_v)\}$, where $\mathbf{F}_v$ are features for utterance $v$ and $\mathbf{S}_v$ is its sentiment label and $\mathbf{E}_v$ is emotion label.\\
    
        \textbf{Output:} Trained ConvXGB Model for Sentiment and Emotion Classification
        \begin{algorithmic}[1]
        \STATE Initialize convolutional network parameters $\Theta_{\text{conv}}$ and XGBoost parameters $\Theta_{\text{xgb}}$.
        \FOR{$e = 1$ to $E$}
            \STATE $\mathbf{C}_v' \leftarrow$ empty list
            \FOR{each utterance $v$ in the training set $D_{\text{Tr}}$}
                \STATE Apply convolution to extract features: \\
                 $\mathbf{C}_{v1, i} = \text{ReLU}\left(\sum_{k=1}^{K} \mathbf{F}_{v, i+k-1} \cdot \mathbf{\Theta}_{\text{conv}, k}\right)$
                \STATE Apply ReLU activation for non-linearity:\\
                $\mathbf{C}_{v1, i} = \max(0, \mathbf{C}_{v1, i})$
                \STATE Apply max pooling for down-sampling:\\
              $\mathbf{C}_{v1, i} = \max(\mathbf{C}_{v1, 2i-1}, \mathbf{C}_{v1, 2i})$
                \STATE Reshape the feature maps for XGBoost input:
            $\mathbf{C}_v' \leftarrow \text{Reshape}(\mathbf{C}_{v1})$
            \ENDFOR
        \ENDFOR
        \STATE Initialize the regularization and tree parameters.
        \FOR {each iteration e = 1 to E}
            \FOR{each tree \(k\) in range(K)}
                \STATE Initialize the leaf scores \(f_k(x)\) for all leaves
                \STATE Do regularization by calculating the loss function and penalizer function to avoid overfitting:
                 $L(\phi) = \sum_{i} \ell\left(\hat{y}_i, y_i\right) + \sum_{k} \Omega\left(f_k\right)$
                \STATE Update tree weights:
                 $\omega_j^{(t)} = -\frac{\sum_{i \in \Gamma_j} \Delta_i}{\sum_{i \in \Gamma_j} \Upsilon_i + \zeta}$
                \STATE Update tree structure:
                $f_j^{(t)} = w_j^{(t)}$ \quad \\
                (Optimal leaf weights)
            \ENDFOR
        \ENDFOR
        \STATE Make predictions using the trained XGBoost model:
         $\mathbf{S}_v \leftarrow \text{XGBoostPredict}(\mathbf{C}_v', \mathbf{\Theta}_{\text{xgb}})$
        \STATE \textbf{return} Final Predictions $\{\mathbf{S}_v \text{ and } \textbf{E}_v, \forall v \in V\}$
    \end{algorithmic}
\end{algorithm}

\textbf{\textit{a) Reshape Layer: }}
To facilitate input into the prediction stage, an internal operation within the reshape layer (refer to \autoref{fig : architecture}) transforms the tensors output from the convolution layers into a vector format, as shown on line 8 in Algorithm~\ref{alg:algo2}. Next, the reshaped features are passed to the classification layer.

\textbf{\textit{b) Classification Layer: }}The functioning of the classification layer is mathematically explained on lines 11-21 in Algorithm~\ref{alg:algo2}. This layer serves primarily for class prediction and leverages the XGBoost algorithm. XGBoost is a tree-based machine learning model that utilizes gradient boosting to sequentially construct an ensemble of decision trees. The number of trees in the ensemble directly influences the model's performance and complexity.

In every cycle, a collection of K trees is used, each tree having \(K_{E}^{i} \mid i \in 1..K\) nodes. The total of the various prediction scores produced by each tree is the final prediction for a particular instance as illustrated in \autoref{eq:predictedscores}:
\begin{equation}
\label{eq:predictedscores}
    \hat{y}_i = \phi\left(x_i\right) = \sum_{k=1}^{K} f_k\left(x_i\right), \quad f_k \in F,
\end{equation}
where, the training set members are denoted by \(x_i\), the associated class labels are denoted by \(y_i\), the leaf score for the \(k^{th}\) tree is represented by \(f_k\), and the set of all \(K\) scores for all trees of classification and regression is represented by \(F\).

Now we define the complexity penalizer function 
 as mathematically illustrated in \autoref{eq:19}.
\begin{equation}
\label{eq:19}
    \Omega(f) = \delta T + \frac{1}{2} \zeta \sum_{j=1}^{T} \omega_j^2
\end{equation}
where, \(T\) is the number of leaves in the tree, \(\omega\) is the weight of each leaf, and \(\delta, \zeta\) are constants governing the regularization degree.

Further, regularization is applied to improve the final result as given in line 15 of Algorithm~\ref{alg:algo2}:
\begin{equation}
\label{eq: eq18}
    L(\phi) = \sum_{i} \ell\left(\hat{y}_i, y_i\right) + \sum_{k} \Omega\left(f_k\right)
\end{equation}

\autoref{eq: eq18} mentions the regularization term \(L(\phi)\) comprising two parts: the complexity penalizer function, which tries to avoid overfitting, and the differentiable loss function \(\ell\), which calculates the difference between the ground truth label \(y_i\) and the prediction \(\hat{y}_i\).

Several regression and classification problems can be handled by gradient boosting. At each phase, the gradient boost loss function is simplified using an extended second-order Taylor expansion to yield a more achievable goal, as follows:
\begin{equation}
\label{eq:gradient-eq}
    \tilde{L}(t) \approx \sum_{i} n \left[ \Delta_i \Phi_i\left(x_i\right) + \frac{1}{2} \Upsilon_i \Phi_i^2\left(x_i\right) \right] + \Omega\left(f_t\right) 
\end{equation}
Further,~\autoref{eq:gradient-eq}, on substitution is further transformed into below form as illustrated in~\autoref{eq:gradient-eq2} :
\begin{equation}
\label{eq:gradient-eq2}
   \tilde{L}(t) = \sum_{j} T \left[ \left(\sum_{i \in \Gamma_j} \Delta_i\right) \omega_j + \frac{1}{2} \left(\sum_{i \in \Gamma_j} \Upsilon_i + \lambda\right) \omega_j^2 \right] + \delta T 
\end{equation}
here, \(\Gamma_j=\{ i \mid \chi\left(x_i\right) = j \}\) denotes the instance set of the leaf \(t\), and 
\(\Delta_i = \frac{\partial \ell\left(\hat{y}_i^{(t-1)}, y_i\right)}{\partial \hat{y}_i^{(t-1)}}\) and
\(\Upsilon_i = \frac{\partial^2 \ell\left(\hat{y}_i^{(t-1)}, y_i\right)}{\left({\partial \hat{y}_i^{(t-1)}}\right)^2}\) are the loss functions of the first and second-order statistics.

Further, \autoref{eq:weight-update} determines the weight \(\omega_j^{(t)}\) for leaf \(j\) at iteration \(t\) based on the ratio of the sum of gradients for the loss function over all samples in the node \(\Gamma_j\) to the sum of second-order gradients and a regularization term \(\zeta\). The weight reflects the leaf node's effectiveness in correcting errors, considering both the loss and regularization.

\begin{equation}
\label{eq:weight-update}
    \omega_j^{(t)} = -\frac{\sum_{i \in \Gamma_j} \Delta_i}{\sum_{i \in \Gamma_j} \Upsilon_i + \zeta},
\end{equation}

The new leaf weight is updated using \autoref{eq:weight-update} iteratively, mentioned in line 16 in Algorithm~\ref{alg:algo2}, ensuring that each new tree focuses on correcting the errors of the previous ones, leading to a progressively more accurate ensemble of decision trees for better feature classification.

\textbf{\textit{ Prediction: }}
The trained XGB model is then employed for sentiment and emotion prediction. Given a set of features for a new input, denoted as {$\mathbf{C}_v'$}, the model predicts the sentiment label $\mathbf{S}_v$ using the XGBoost prediction function as illustrated on lines 20-21 in Algorithm~\ref{alg:algo2}.
The resulting sentiment predictions provide valuable insights into the sentiment conveyed by the multimodal input. Similarly, for Emotion prediction the XGB model will be given the feature tensor as input denoted by {$\mathbf{C}_v'$}, the model predicts the emotion label $\mathbf{E}_v$ using the XGB predict function.

% \begin{algorithm}
%     \caption{\textsc{EmoXGB: Enhanced Emotion Prediction}}
%     \begin{algorithmic}
%         \vspace{1mm}
%         \STATE \textbf{Input:} $D_{\text{Train}} = \{(\mathbf{X}_v, \mathbf{E}_v)\}$ \COMMENT{Training dataset}
%         \STATE \textbf{Output:} Trained EmoXGB Model
%         \end{algorithmic}
%         \begin{algorithmic}[1]
%         \vspace{1mm}
%         \FOR{$e = 1$ to $E$}
%             \FOR{each utterance $v$ in $D_{\text{Train}}$}
%                 \STATE Apply convolution and pooling to extract features: $\mathbf{C}_v' \leftarrow \text{Reshape}(\max(0, \sum_{k=1}^{K} \mathbf{X}_{v, i+k-1} \cdot \mathbf{\Theta}_{\text{conv}, k}))$
%             \ENDFOR
%         \ENDFOR
%             \STATE Train XGBoost on reshaped features using emotion labels
%         \FOR{each iteration e = 1 to E}
%             \FOR{each tree $k$ in range (K)}
%                 \STATE Initializing leaf scores and calculating loss function.
%                 \STATE Updating tree leaf weights and use penalising function to decrease errors in each iteration.
%             \ENDFOR
%             \STATE $\mathbf{\Theta}_{\text{xgb}} \leftarrow \{\mathbf{F}^{(t)}, \forall t\}$ \COMMENT{Trained XGBoost Model}
%         \ENDFOR
%         \STATE Make predictions using trained XGBoost model: $\mathbf{E}_v \leftarrow \text{XGBoostPredict}(\mathbf{C}_v', \mathbf{\Theta}_{\text{xgb}})$
%         \vspace{1mm}
%         \STATE\textbf{return} Emotion Predictions $\{\mathbf{E}_v, \forall v \in V\}$
%     \end{algorithmic}
% \end{algorithm}

% \newpage
\section{\MakeUppercase{Experimental Evaluations}}
\label{section:experimental-evaluation}
This section exhibits the validity of our technique by first introducing the experimental setup and then demonstrating the results of the experiment.
\subsection{Experimental Setup}
The following section summarizes the extensive datasets that have been employed for the experimentations. We next go over preliminary techniques for comparison, followed by the adopted evaluation metrics.
\subsubsection{Datasets}

In the process of evaluating the accuracy of our proposed approach for multimodal sentiment and emotion analysis, we selected and employed three benchmark datasets. To evaluate the efficiency of the sentiment analysis task, we leveraged the well-regarded CMU-MOSI and CMU-MOSEI datasets. Furthermore, we utilize the IEMOCAP dataset for emotion prediction. The subsequent sections provide an in-depth explanation of the datasets shown in \autoref{tbl:datasets}.

\begin{table}[width=0.99\linewidth,cols=5,pos=h]
 \caption{Comparative Analysis of Incorporated Datasets: We assess our proposed technique on three publicly available multimodal datasets, namely CMU-MOSI, CMU-MOSEI, and IEMOCAP. The following table presents the metadata about the datasets and the information about the content of which the datasets are comprised. } \label{tbl:datasets}
    \centering
    \begin{tabular}{cccccccccc}
    \toprule
        \textbf{Dataset } & \textbf{Videos} & \textbf{Utterances} & \textbf{Speakers} & \textbf{Language} & \textbf{Source} & \textbf{Topics}\\
        \midrule
          CMU-MOSI & 93 & 2199 & 89 & Multiple & YouTube & Movie reviews \\
        CMU-MOSEI  & 5000 & 23453 & 1000 & Multiple & YouTube & {Reviews \& debate} \\
       IEMOCAP  & 100 & 1271 & 10 & English & USC Viterbi &  Improvisations \& scripted scenarios \\
       \bottomrule
    \end{tabular}
   
\end{table}
\noindent \textbf{\textit{(a) CMU-MOSI:}}
We incorporate the CMU-MOSI dataset in our research. The aforementioned data set was first published by Amir Zadeh et al \citep{zadeh2016mosi}. It stands as a prominent benchmark, particularly well-suited for evaluating the performance of fusion networks in the challenging task of sentiment intensity prediction. Comprising an array of YouTube video blogs (vlogs), this dataset captures the diverse expressions of speakers articulating their opinions across various topics. In its entirety, it consists of 2,199 curated utterance-video segments sourced from 93 videos, each featuring a unique narrator. This dataset is distinctive due to its rigorous manual annotation process, where each segment is assigned a real-number score in a range from \(-3\) to \(+3\). This score is a measure of the relative strength of emotions—negative sentiments have values below zero and positive sentiments have values more than zero.
\\

\noindent \textbf{\textit{(b) CMU-MOSEI:}}
Building upon the foundation of CMU-MOSI \citep{bagher-zadeh-etal-2018-multimodal}, the CMU-MOSEI dataset emerges as an enriched counterpart, both in terms of sample size and speaker diversity. With an expanded set of samples totaling 23,453 video segments, CMU-MOSEI captures a broader spectrum of human experiences and opinions. These segments undergo manual annotation, maintaining the real-number score convention for sentiment intensity. This expansive dataset spans 5,000 videos, engaging 1,000 distinct speakers and spanning 250 unique topics. CMU-MOSEI thus provides a comprehensive and diverse set of multimedia instances for the exploration and evaluation of multimodal sentiment analysis.
\\

\noindent    \textbf{\textit{(c) IEMOCAP:}}
For multimodal emotion recognition, the Interactive Emotional Dyadic Motion Capture (IEMOCAP)~\citep{Busso2008IEMOCAPIE} dataset offers a unique and rich resource. Comprising a total of 12 hours of audio-visual data, IEMOCAP captures dialogues between 10 actors engaged in both scripted and improvised conversations. Following data collection, the audio-visual content is segmented into smaller utterances, each lasting between 3 to 15 seconds. The uniqueness of IEMOCAP lies in its detailed labeling process. Each utterance undergoes evaluation by 3-4 assessors, using a 10-option scale encompassing a wide range of emotions. For our analysis, we focus on four emotions—anger, excitement (happiness), neutrality, and sadness, keeping consistent with prior research and representing emotions where at least 2 experts were in agreement. This stringent labeling ensures a robust and reliable dataset, aligning with established practices in emotion research.
% \begin{table*}[htp]

\subsubsection{Compared Methods}
To assess the efficiency of our proposed technique, we compared it to other recent sentiment and emotion analysis methods as discussed below.
\vspace{2mm} \newline
\textbf{\textit{(a) Multimodal Fusion Network (MFN):}}
The Multimodal Fusion Network (MFN)~\citep{mfn} established the feasibility of early fusion in multimodal sentiment analysis. Its key innovation was the direct concatenation of feature vectors extracted from disparate modalities (e.g., text, audio, video) into a single representation. This approach bypassed intermediate feature-level fusion and fed the combined vector directly into a sentiment classifier. While seemingly simplistic, MFN demonstrated remarkable effectiveness in capturing cross-modal correlations and identifying basic sentiment patterns across diverse information sources. Its success laid to the foundation of early fusion techniques in multimodal learning tasks.
\vspace{2mm}\newline
\textbf{\textit{(b) Tensor Fusion Network (TFN):}} The Tensor Fusion Network (TFN)~\citep{tfn} builds upon the success of Multimodal Fusion Network(MFN) by refining its early fusion approach. Unlike MFN's simple concatenation, TFN leverages the expressive power of tensor products. For each modality, TFN employs separate subnetworks to extract modality-specific feature vectors. These vectors are then combined through an outer product, which generates a higher-order tensor capturing both inter- and intramodal interactions. This multimodal tensor undergoes subsequent transformations to learn complex fusion patterns beyond basic correlations. Notably, the tensor product operation is parameter-free, reducing overfitting risks and potentially facilitating the interpretability of the learned multimodal representation. In essence, TFN elevates early fusion beyond mere feature aggregation, leading to a richer and more specific understanding of sentiment across modalities.
\vspace{2mm}\newline
\textbf{\textit{(c) Multi-attention Recurrent Network (MARN): }}
The MARN \citep{marn} tackles multimodal sentiment and emotion analysis by first capturing the essence of each modality (text, audio, video) through separate subnetworks. These subnetworks extract features specific to each modality, like word meanings, vocal tone, and facial expressions. A clever ``multi-attention'' mechanism then analyzes these features within each modality and across modalities over time, pinpointing which aspects are most relevant to the overall sentiment or emotion being conveyed. This dynamic attention dynamically weights the features, creating a systematic understanding of how different modalities interact and contribute to the emotional message. Finally, a ``Long-short Term Hybrid Memory'' component carefully stores and integrates this rich information, capturing both fleeting emotions and deeper sentiments across the entire communication sequence. The resulting comprehensive representation is then fed into a classifier to accurately pinpoint the overall sentiment or emotion expressed.
\vspace{2mm}\newline
\textbf{\textit{(d) Modality-Invariant and Specific Subspaces (MISA):}}
Complex fusion methods in multimodal tasks can struggle with morphological gaps between different modalities. To address this challenge, Hazarika et al.~\citep{misa}, proposed MISA, a novel framework that leverages modal subspaces to enhance the fusion process. The core contribution of MISA lies in its modal representation learning stage, which precedes the actual fusion step.
Following feature extraction for each modality (audio, visual, and text), MISA projects each modality into two distinct subspaces. The first subspace is modality-invariant, aiming to capture the commonalities between modalities by minimizing the heterogeneity gap through a distribution similarity constraint. Conversely, the second subspace is modality-specific, focusing on learning unique feature information specific to each modality. After subspace projection, a transformer-based self-attention mechanism is employed to concatenate all six transformed modal vectors. This combined representation is then fed into simple feed-forward layers for prediction. Notably, by exploring the feature space through subspace learning, MISA reduces the reliance on complex fusion mechanisms, potentially leading to improved performance.
\vspace{2mm}\newline
\textbf{\textit{(e) Speaker-Independent Multimodal Representation:}}
The SIMR framework, as proposed by \citep{simr}, strategically partitions nonverbal data into distinct components, namely style encoding and content representation. This deliberate separation serves to mitigate the impact of personalized acoustic and visual features. Simultaneously, the framework adeptly uncovers both compatible and incompatible cross-modal interactions through the integration of an enhanced Transformer module. By dissecting nonverbal inputs into style encoding and content representation, the framework leverages informative cross-modal correlations. Unlike conventional transformer-based approaches that primarily focus on discovering compatible cross-modal interactions, our methodology goes a step further. It not only identifies compatible interactions but also pays due attention to incompatible ones, achieved through the incorporation of an enhanced cross-modal transformer module. This systematic approach ensures a more comprehensive understanding of the interplay between modalities, enhancing the model's ability to handle speaker-independent multimodal representation effectively.
\vspace{2mm}\newline
\noindent \textbf{\textit{(f) Multi-level Correlation Mining Framework (MCMF):}}
This work introduces a novel approach to multimodal sentiment analysis, addressing challenges related to feature fusion and co-learning. Their proposed method \citep{mcmf} incorporates a multilevel correlation mining framework and a self-supervised label generation module. Leveraging unimodal features fusion and a linguistics-guided transformer, the model effectively integrates low and high-level correlation information. A multi-task learning framework facilitates co-learning, while the self-supervised label generation module overcomes the lack of unimodal labels. The study's key contributions include enhancing fusion through unimodal features fusion, addressing multimodal complexity with linguistics-guided transformers, and providing a comprehensive solution to co-learning challenges. The results demonstrate the model's effectiveness in multimodal sentiment analysis.
\vspace{2mm}\newline
\noindent \textbf{\textit{(g) Multimodal Transformer (MulT):}}
The Multimodal Transformer (MulT)~\citep{tsai-etal-2019-multimodal}, recognizes emotions by processing multimodal data, including language, facial movements, and audio behaviors, without explicit alignment.  capture associated crossmodal information, it employs a directional pairwise crossmodal attention mechanism to handle interactions across several modalities and time steps. Our method varies from MulT in that it includes an Intermodal Contextual Interaction Module that dynamically assigns weights to each modality's representation and a harmonic optimization algorithm to address data redundancy. MulT focuses on handling non-alignment of data and long-range dependencies. This difference emphasizes how we optimize feature contributions and fusion efficiency, while MulT focuses on attention processes.
\vspace{2mm}\newline
\noindent \textbf{\textit{(h) Multimodal End-to-End Sparse Model (MESM):}}
In order to improve emotion recognition, the Multimodal End-to-End Sparse Model (MESM)~\citep{DBLP:journals/corr/abs-2103-09666}, combines feature extraction and model training into a single end-to-end procedure. The standard two-phase pipeline has drawbacks that MESM solves. Specifically, its fixed features cannot be adjusted to suit varied workloads. Whereas, with MESM, the performance is maintained at a lower computational overhead with the introduction of a sparse cross-modal attention mechanism and the reorganization of datasets for end-to-end training. Tests reveal that MESM outperforms cutting-edge models built on the two-phase pipeline. In contrast to MESM, GCM-Net emphasizes feature contribution optimization by dynamic weighting and effective feature selection, demonstrating distinct approaches to multimodal emotion identification problems.

\subsubsection{Evaluation Metrics}

Our proposed model GCM-Net, aims to extract sentiment and emotions from user videos by analyzing multiple modalities, including visual, textual, and audio modalities. To gauge the effectiveness of this method, it's crucial to evaluate their performance using appropriate metrics. This article focuses on four widely employed metrics: accuracy, precision, recall, and F1-score.

The precision parameter measures the proportion of correctly predicted instances for a specific label within a binary classification task, where, $label \in \{positive, negative \}$. It is denoted as the ratio of the number of correctly predicted instances of the specific label ($label \in {positive,negative}$) to the total number of instances predicted with that label, as shown in \autoref{eq:eq23}.
\begin{equation}
 Precision_{label} =  \frac{True\_ Predicted_{label}  }{Total\_Predicted _{label}}
\label{eq:eq23}
\end{equation}

Recall, also termed sensitivity, measures the proportion of true positives within the total number of actual positive cases. It reflects the model's ability to correctly identify all relevant instances belonging to a specific class. Mathematically, recall is calculated as: \autoref{eq:eq24}.
\begin{equation}
 Recall_{label} = \frac{ True\_Predicted_{label}  }{Total_{label}}
\label{eq:eq24}
\end{equation}

\begin{equation}
 F1-score_{label} = \frac{2 \times Precision_{label} \times Recall_{label} } { Precision_{label} + Recall_{label} }
\label{eq:eq25}
\end{equation}

The F1-score is a widely used metric that combines precision and recall into a single, harmonic mean value. This metric aims to provide a balanced evaluation of a model's performance by considering both its ability to correctly identify positive instances (precision) and its ability to capture all relevant positive instances (recall). F1-score can be mathematically calculated as shown in \autoref{eq:eq25}.

\subsection{Experimental Results}
This section summarizes the experimental outcomes derived from existing methods against our proposed method. We evaluate the proposed technique by comparing its effectiveness with existing methods on varied datasets, visualizing the recommendations through qualitative analysis, analyzing performance gain, and identifying the sensitivity of various parameters. 

\subsubsection{ Effectiveness Comparisons:}
We compare the performance of GCM-Net with the existing methods on three benchmark datasets. The performance over sentiment analysis is evaluated on CMU-MOSI and CMU-MOSEI datasets, whereas the IEMOCAP dataset is leveraged for the emotion analysis task.
\vspace{2 mm}

\noindent \textbf{\textit{(a) Performance on CMU-MOSI and CMU-MOSEI for Sentiment Analysis:}}
In assessing the efficacy of our proposed model for multimodal sentiment analysis, we conducted a comparative analysis with existing methods, as per prior research practices. 
\begin{table}[width=0.85\linewidth,cols=5,pos=h]
\caption{Comparison of our GCM-Net with Existing Models on Two-Class Accuracy }\label{tbl:combined_model_comparison}
\begin{tabular*}{\tblwidth}{@{} LLLLLL@{} }
\hline
\multirow{2}{*}{\textbf{Model}} & \multicolumn{2}{c}{\textbf{MOSI}} & \multicolumn{2}{c}{\textbf{MOSEI}} \\
& \textbf{Accuracy} & \textbf{F1-score} & \textbf{Accuracy} & \textbf{F1-score} \\
\hline
TFN \citep{tfn}& 0.7460 & 0.7450 & 0.7560 & 0.7550 \\
MARN \citep{marn}& 0.7710& 0.7700& 0.7930 & 0.7780\\
MFN \citep{mfn} & 0.7740 & 0.7740 & 0.7990 & 0.7910 \\
% RAVEN \citep{raven}& 0.7800 & 0.7660 & 0.7910 & 0.7951 \\
MISA \citep{misa}& 0.8180 & 0.8187 & 0.8360 & 0.8380 \\
% SKEAFN \citep{skeafn}& 0.8734 &0.8734& 0.8707 & 0.8719 \\
SIMR  \citep{simr}& 0.8610 &0.8610 & 0.8320 & 0.8320 \\
MCMF \citep{mcmf}&0.8843 & 0.8843 & 0.8616  & 0.8588 \\
\textbf{GCM-Net} & \textbf{0.9266} & \textbf{0.9444} & \textbf{0.8657} & \textbf{0.8923}\\
\hline
\end{tabular*}
\end{table}

\begin{figure}[h]
     \centering
    \includegraphics[width=0.6\linewidth]{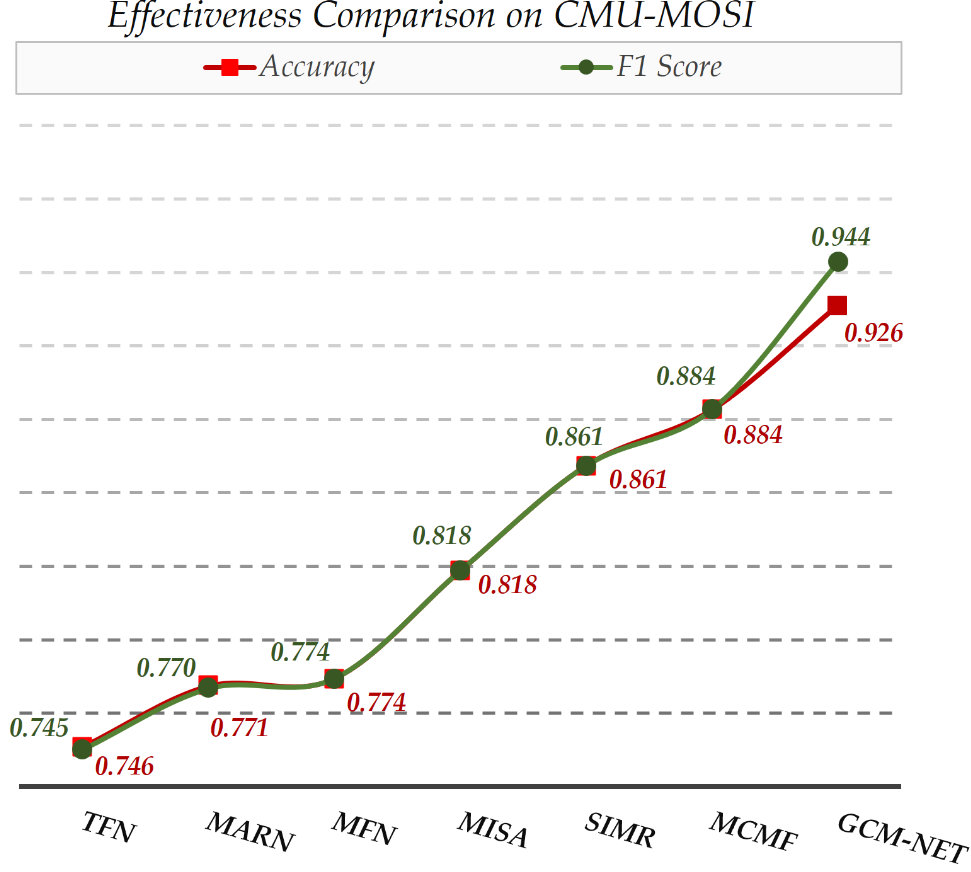}
    \caption{Performance of Considered Models on CMU-MOSI for Sentiment Analysis}
    \label{fig:mosi-graph}
\end{figure}

The performance evaluation encompasses metrics such as accuracy and F1-score on the CMU-MOSI dataset, and the results are presented in \autoref{tbl:combined_model_comparison}. The tabulated data clearly illustrates that our model surpasses previous state-of-the-art approaches on the specified dataset by a significant margin. Specifically, GCM-Net exhibits a noteworthy improvement of 18.06\% and 19.94\% in accuracy and F1-score, respectively, over TFN. This enhancement is attributed to the utilization of graph-based feature reconstruction, which considers temporal context and interrelations with neighboring features. 

In contrast, TFN relies solely on late fusion, limiting its capacity to learn intricate inter-modality associativity. Furthermore, in comparison to early fusion techniques like MFN, our model demonstrates a performance boost of 15.26\% and 17.04\% in accuracy and F1-score. Overcoming the challenge of feature redundancy in MFN, our model employs a metaheuristic algorithm for feature selection, focusing only on crucial features for sentiment prediction. 

Considering the MCMF model, which leverages correlation information between modalities at various levels, our model showcases superiority with improvements of 4.23\% and 6.01\% in accuracy and F1-score. Additionally, our model exhibits advancements of 10.86\% and 12.57\% in accuracy and F1-score, respectively, over MISA. The increasing accuracy trend is illustrated in \autoref{fig:mosi-graph}. These significant improvements in both metrics across various existing methods underscore the competitive edge and effectiveness of our proposed technique.
    \\

% \noindent\textbf{\textit{(b) Performance on CMU-MOSEI for Sentiment Analysis:}}

\begin{figure}[htp]
     \centering
    \includegraphics[width=0.6\linewidth]{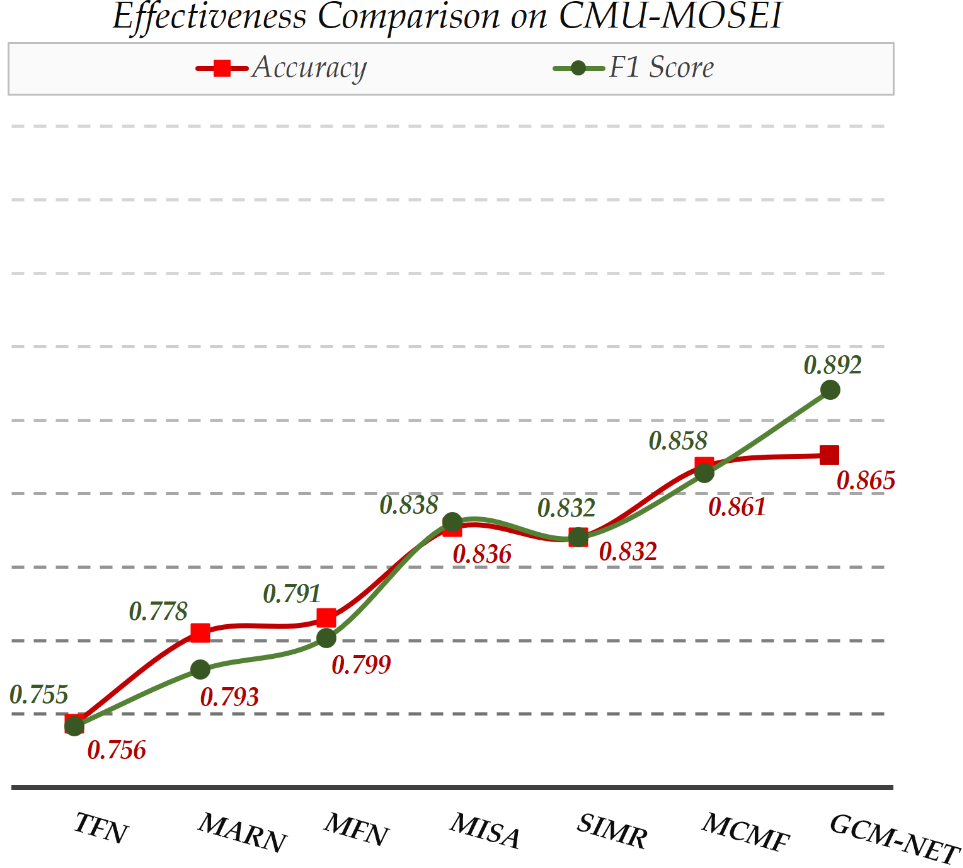}
    \caption{Performance of Considered Models on CMU-MOSEI for Sentiment Analysis}
    \label{fig:mosei-graph}
\end{figure}
Continuing our comprehensive evaluation, we extend our model comparison to the CMU-MOSEI dataset. The outcomes, detailed in \autoref{tbl:combined_model_comparison}, reveal the consistent superiority of our proposed model over existing benchmarks.

Notably, GCM-Net achieves an impressive improvement of 11.37\% in accuracy and 13.73\% in F1-score over TFN. This notable advancement can be attributed to our model's adept utilization of graph-based feature recalibration, capturing temporal context and intermodal relationships effectively. Compared with early fusion techniques such as MFN, our model demonstrates a remarkable boost of 5.97\% in accuracy and 10.13\% in the F1-score. Addressing the challenge of feature redundancy, our model employs a metaheuristic algorithm for feature selection, enhancing its discernment of critical features for sentiment prediction. Considering the SIMR model, which exploits correlation information between modalities, our model showcases a substantial lead with improvements of 2.67\% in accuracy and 6.03\% in the F1-score. Additionally, our model exhibits advancements of 1.97\% in accuracy and 5.43\% in F1-score over MISA. As shown in \autoref{fig:mosei-graph}, consistent improvements over these datasets underscore the robustness and effectiveness of our proposed model in handling diverse multimodal sentiment analysis tasks.
\vspace{4 mm}\newline
\noindent\textbf{\textit{(c) Performance on IEMOCAP for Emotion Analysis:}}
To justify the effectiveness of our suggested approach, we examine its results on the IEMOCAP dataset, which is well-known as a standard dataset for complex emotion identification problems.

\begin{table}[width=12.5 cm,cols=5,pos=h]
\caption{Comparison of GCM-Net with Existing Models for IEMOCAP Dataset }\label{tbl:combined_model_comparison_IEMOCAP}
\begin{tabular*}{\tblwidth}{@{} LLLLL@{} }
\toprule
\textbf{Models} & \textbf{Accuracy} & \textbf{F1-score} \\
\midrule
LF-LSTM & 0.7180 & 0.4950 &\\
LF-TRANS & 0.7880& 0.5030 \\
EmoEmbs \citep{DBLP:journals/corr/abs-2009-09629} & 0.7720 & 0.4980  \\
MulT \citep{tsai-etal-2019-multimodal}& 0.7760 & 0.5690 \\
CMHA  \citep{9693238}& 0.8420 & 0.5610  \\
MESM \citep{DBLP:journals/corr/abs-2103-09666}&0.8440 & 0.5740  \\
FE2E \citep{dai-etal-2021-multimodal}&0.8450 & 0.5880  \\
\textbf{GCM-Net} & \textbf{0.8566} & \textbf{0.7269}\\
\hline
\end{tabular*}
\end{table}
As illustrated our model GCM-Net stands out, achieving remarkable advancements over prior methods. Compared to LF-LSTM, the prevailing benchmark, GCM-Net exhibits a significant stride of 13.86\% in average accuracy and 23.19\% in F1-score. This dominance extends to other competitive models like LF-TRANS, EmoEmbs, and MulT, demonstrating improvements of 6.86\%, 8.46\%, and 8.06\% in accuracy respectively, alongside notable F1-score gains. Also, the popular approaches like CMHA, MESM, and FE2E, are outperformed by GCM-Net by 1.46\%, 1.26\%, and 1.16\% in accuracy respectively, establishing a remarkable 15.29\% lead in F1-score over FE2E. This comparison of performances is tabulated in \autoref{tbl:combined_model_comparison_IEMOCAP}.
\begin{figure}[htp]
    \centering
    \includegraphics[width=0.6\linewidth ]{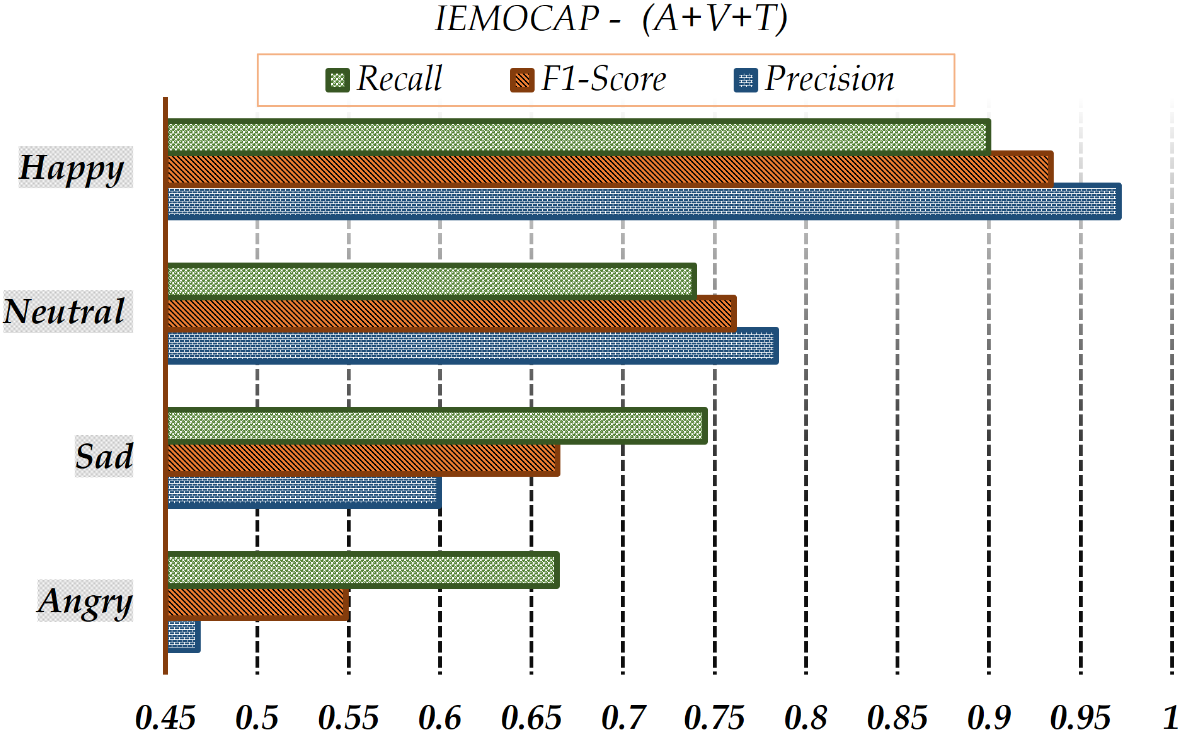}
    \caption{Label-specific Prediction Results on IEMOCAP}
    \label{fig:IEMOCAP}
\end{figure}
The emotion prediction results highlight the effectiveness of our multimodal approach in identifying subtle inherent emotions in video data. Our chosen benchmark dataset, IEMOCAP~\citep{Busso2008IEMOCAPIE}, exhibits a class imbalance. While the dataset encompasses utterances categorized into four emotions, the distribution is uneven. Specifically, the number of utterances labeled as ``Happy'' significantly outnumbers those labeled as ``Angry''. \autoref{fig:IEMOCAP} provides recall, F1-score, and precision for each emotion category: Happy, Angry, Sad, and Neutral, all presented in decimal form for clarity.

We emphasize the tri-modal combination (A + T + V), which integrates acoustic, textual, and visual modalities. This comprehensive approach facilitates a deeper understanding of emotional expressions, leveraging the synergies between audio, textual, and visual modalities as shown in \autoref{fig:IEMOCAP}. This strategic integration enhances our model's ability to interpret and decode the inherent emotion of the utterances, showcasing the precision and depth of our analysis.

% \paragraph{\textbf{\textit{Effectiveness Comparison on IEMOCAP dataset:}}}
% \subparagraph{}

% \begin{figure}
%     \centering
%     \includegraphics[width=0.98\linewidth]{graph-iemocap.png}
%     \caption{Performance of considered models on IEMOCAP for Emotion Prediction}
%     \label{fig:graph-iemocap}
% \end{figure}

\subsubsection{ Performance Gain Analysis on CMU-MOSI for Sentiment Analysis:}
In this section, we analyze the performance gain of the proposed method. We first examine the performance of GCM-Net with different modalities
combinations followed by different attention mechanisms.
\begin{table}[width=.7\linewidth,cols=4,pos=h]
\caption{Ablation Study of GCM-Net on CMU-MOSI for Sentiment Analysis for different Modality Combinations}\label{tbl2}
\begin{tabular*}{\tblwidth}{@{} LLLL @{} }
\toprule
\textbf{Modality} & \textbf{Accuracy} & \textbf{F1-score} & \textbf{Recall}\\
\midrule
V & 0.8325 & 0.8976 & 0.9339\\
A & 0.8545 & 0.8981 & 0.8150\\
T & 0.8580 & 0.9167 & 0.9933\\
A + V & 0.8413 & 0.8967 & 0.8756\\
A + T & 0.8915 & 0.9349 & 0.9899\\
T + V & 0.8968 & 0.9229 & 0.8777\\
A + T + V & 0.9266 & 0.9311 & 0.8857\\
\bottomrule
\end{tabular*}
\end{table}
\newline
\noindent \textbf{\textit{(a) Modality Combinations: }}
As elaborated in \autoref{tbl2}, we explain the ablation studies done based on the combinations of various modalities taken into consideration. The results derived from the bi-modal combinations indicate that opting for the text-acoustic combination is preferable over other choices, as it leads to improved performance. Ultimately, we conduct experiments involving tri-modal inputs and gain a notably enhanced performance. This observation underscores the significance of employing a combination that integrates all three modalities, emphasizing its superiority in achieving improved results.

\vspace{2 mm}
\noindent  \textbf{\textit{(b) Feature Recaliberation using Graph Neural Network: }}
We initiate by presenting the results of incorporating the Graph Neural Network in \autoref{tbl:enhancement} that illustrate the impact of Feature Recalibration using Graph Neural Network in our architecture compared to a scenario without its implementation and different combinations of all the modules for drawing out better conclusions. The performance gain is expressed as the difference between the two approaches.
   %  \begin{table}[width=.9\linewidth,cols=4,pos=h]
   %  \caption{With and Without Feature Recalibration}\label{tbl:tbl5}
   %  \begin{tabular*}{\tblwidth}{@{} LLLL@{} }
   %  \toprule
   % \textbf{ Feature Recalibration} & \textbf{Accuracy}\\
   %  \midrule
   %  With Feature Recalibration & 0.9266 \\
   %  Without Feature Recalibration & 0.8560\\
   %  \bottomrule
   %  \end{tabular*}
   %  \end{table}
Feature recaliberation significantly enhances accuracy, demonstrating its effectiveness in capturing complex relationships in multimodal data.
\begin{table}
[width=.7\linewidth,cols=2,pos=h]
\caption{Ablation Study of GCM-Net on CMU-MOSI for Sentiment Analysis with Combination of Modules in Architecture}
    
\label{tbl:enhancement}
\centering
\begin{tabular*}{\tblwidth}{LL}
\toprule
\textbf{Modules} & \textbf{Accuracy}\\
\midrule
FRE & 0.8210 \\
ICIM & 0.8130 \\
HOA & 0.8348 \\
FRE + ICIM & 0.8320\\
FRE + SCA & 0.8743\\
SCA + ICIM & 0.8560\\
FRE + SCA + ICIM & 0.9266\\
\bottomrule
\end{tabular*}
\end{table}

\noindent \textbf{    \textit{(c) Contextual Enrichment by Attention Mechanism: }}
Next, we discuss the performance gain achieved by incorporating Cross-Modal Attention mechanisms into our model as illustrated in \autoref{tbl:enhancement}. We study the attention values to better understand the proposed architecture's behavior while learning. The outcomes that were achieved show that by applying distinct weights across separate modalities and contextual utterances, the model accurately predicts the labels of the experimented utterances. 
    % \begin{table}[width=.9\linewidth,cols=4,pos=htp]
    % \caption{With and Without Cross-Modal Attention}\label{tbl:tbl6}
    % \begin{tabular*}{\tblwidth}{@{} LLLL@{} }
    % \toprule
    % \textbf{Cross-Modal Attention} & \textbf{Accuracy} \\
    % \midrule
    % With Cross-Modal Attention  & 0.9266 \\
    % Without Cross-Modal Attention  & 0.8743 \\
    % \bottomrule
    % \end{tabular*}
    % \end{table}
    The utilization of Cross-Modal Attention contributes to a notable improvement in accuracy, emphasizing its role in capturing relevant features across diverse modalities.
\vspace{2 mm}\newline
\noindent\textbf{\textit{(d) Feature Selection using A\(\beta\)HC Integrated HOA: }}
Finally, we investigate the performance gain resulting from the integration of A\(\beta\)HC encased HOA (Semantic Correspondence Attention) as illustrated in \autoref{tbl:enhancement}.
   %  \begin{table}[width=.9\linewidth,cols=4,pos=htp]
   %  \caption{With and Without HOA}\label{tbl:tbl7}
   %  \begin{tabular*}{\tblwidth}{@{} LLLL@{} }
   %  \toprule
   % \textbf{ A\(\beta\)HC Encased HOA} & \textbf{Accuracy} \\
   %  \midrule
   %  With A\(\beta\)HC Encased HOA  & 0.9266 \\
   %  Without A\(\beta\)HC Encased HOA  & 0.8320 \\
   %  \bottomrule
   %  \end{tabular*}
   %  \end{table}
The integration of A\(\beta\)HC encased HOA results in a substantial accuracy improvement, highlighting its effectiveness in capturing semantic correspondences and enhancing feature representations.

In the upcoming section, we will discuss about qualitative analysis carried out for our proposed model.
\subsubsection{Qualitative Analysis}
\begin{figure*}[h]
    \centering
    \includegraphics[width=\textwidth]{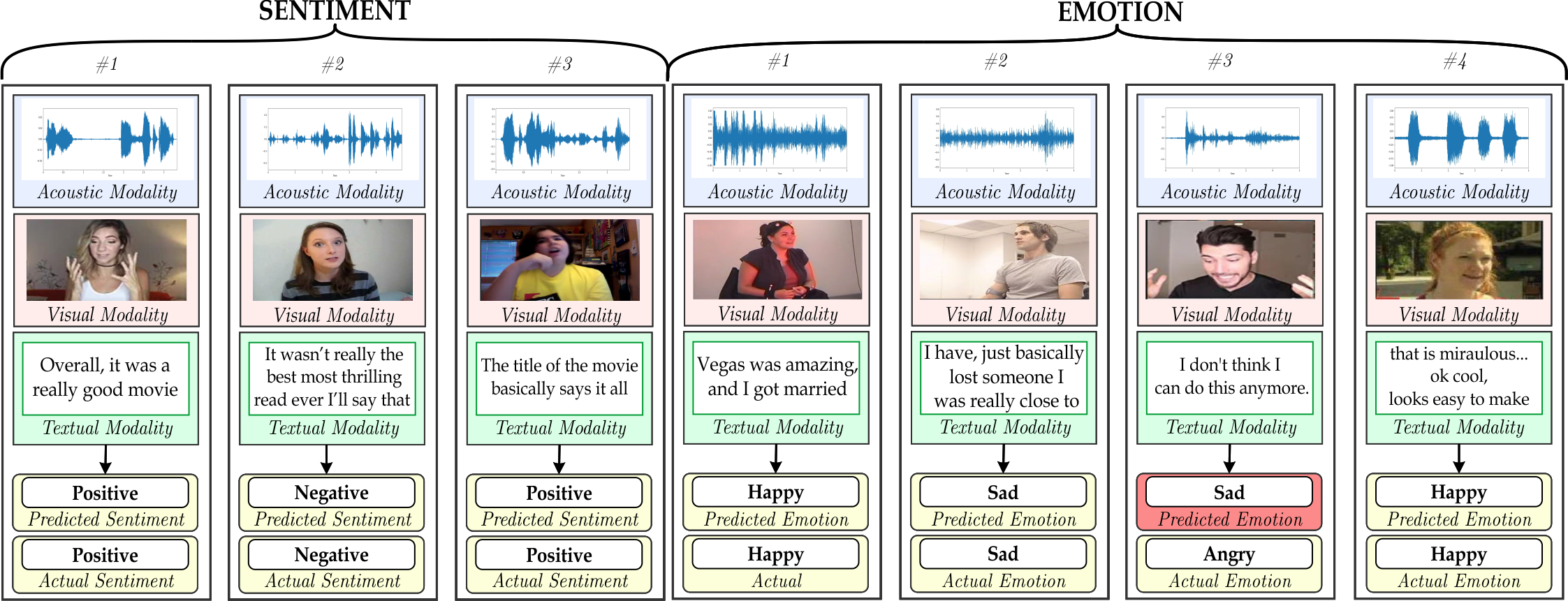}
    \caption{Qualitative Results on CMU-MOSI and IEMOCAP:This figure illustrates qualitative results on corresponding datasets. We descriptively show modality specific data points by portraying its audio plot, visual frames and textual transcript. These are futher classified by GCM-Net for sentiment analysis on CMU-MOSI. For Emotion Analysis the utterances of IEMOCAP are classified by GCM-Net with respect to four-class labels - Happy, Angry, Sad and Neutral.}
    \label{fig : QAnalysis}
\end{figure*}

Multimodal sentiment analysis and emotion analysis approaches typically evaluate effectiveness by accurately classifying utterance data into corresponding sentiment or emotion classes. This section presents a qualitative analysis to assess our proposed model's performance in classifying data points within both sentiment and emotion categories. We showcase the qualitative results on the CMU-MOSI dataset for the sentiment analysis task and on the IEMOCAP dataset for the emotion analysis task. In~\autoref{fig : QAnalysis}, we first show the individual modality elements from the utterances illustrated by the audio plot, associated video frames, and its corresponding textual transcript. We perform a binary classification of sentiments through positive and negative labels and compare the inherent and the predicted sentiment class to evaluate the distinctive capability of GCM-Net. For emotion classification, we assess the emotions by portraying in a similar manner by classifying them into one among four emotion classes namely- Happy, Angry, Sad, and Neutral. The exhibited utterances were chosen from the test data, with yellow blocks representing accurately predicted data labels and red showing uncertainty in prediction. Here, accurately refers to the predicted sentiments that match the ground-truth labels of the utterances, and ambiguous refers to those that do not align with ground-truth labels. As illustrated in~\autoref{fig : QAnalysis}, GCM-Net performs excellently in the binary sentiment classification of the stated utterances. Although, in the IEMOCAP dataset, we encountered an instance where an ``Angry'' emotion was expressed through text conveying dullness and facial expressions appearing upset, where our model categorized it as ``Sad''. This case highlights the potential of complexities in multimodal emotion analysis~\citep{fu2024hybrid}, where the model can sometimes deviate from the ground-truth class labels leading to ambiguity in prediction. Ultimately, in an overall sense, we see that GCM-Net shows better results with respect to considered tasks in comparison to the existing methods. Its ability to grasp inter-modal contextual information and optimal feature selection makes it ideal for combined sentiment and emotion classification.
\subsubsection{Parameter Study}
This section explores various parameters in our proposed model. We examine the impact of four key parameters: the threshold for creating the similarity matrix ($K_{threshold}$), the dropouts in Bi-GRU layers, the number of agents and iterations in the Harmonic optimization Algorithm (HOA) for feature selection (represented as $M_{agents}$ and $M_{iterations}$ respectively), and specific parameters in ConvXGB. For $K_{threshold}$, the optimal value yielding the best results was found to be 0.7. Values below 0.7 were disregarded as they provided insufficient similarity for relevant details, resulting in reduced data redundancy. Higher values led to a sparse matrix with limited utility.
We integrated modality-specific Bi-GRU with 300 neurons each, followed by dense layers of 100 neurons (MOSI) and 128 neurons (MOSEI and IEMOCAP) to standardize the input dimensions across modalities. Dropout regularization was optimally set to 0.7 (MOSI \& IEMOCAP) \& 0.5 (MOSEI) for overall regularization, and 0.5 for Bi-GRU layers.

Further, the value of $M_{agents}$ was set to 4, and value of $M_{iterations}$ to 100 to leverage the local search algorithm extensively for optimal feature selection. Lower values for $M_{agents}$ and the $M_{iterations}$ parameters were deemed insufficient for meaningful comparisons and obtaining an optimized feature set.
Finally, Extreme Gradient Boost parameters, particularly alpha and gamma, values of 0.6 and 0.5 respectively were chosen for the emotion classification task. This decision was influenced by the class imbalance in the dataset.

\section{\MakeUppercase{Conclusion}}
\label{section:conclusion}

In this paper, we introduce GCM-Net, a novel unified framework for multimodal sentiment and emotion analysis. We aimed to develop a robust framework that effectively captured the distinct characteristics across modalities. GCM-Net hierarchically maximizes contextual information through mutual multimodal learning and is validated on three benchmark datasets. GCM-Net's robust performance is facilitated by contextually refined embeddings and attention mechanisms assisted by the metaheuristic algorithm used for harmonic optimization of the feature set. Recognizing limitations in previous works, our approach addressed issues of ineffective modality fusion, intermodal contextual congruity, and suboptimal feature space optimization. This showed significant performance improvements, emphasizing its potential for real-world applications that require in-depth analysis. GCM-Net contributes to the advancement by offering a solution that can assess complex sentimental and emotional notions in social media content. This ability to interpret diverse emotions and sentiments enhances our understanding of social media dynamics and user behavior.

\bibliographystyle{cas-model2-names}

\bibliography{main}
\end{document}